\documentclass[10pt,twocolumn,letterpaper]{article}
\usepackage{titling}

\usepackage[pagenumbers]{cvpr} 

\usepackage{graphicx}
\usepackage{amsmath}
\usepackage{amssymb}
\usepackage{booktabs}
\usepackage{multirow}
\usepackage{tcolorbox}
\usepackage{adjustbox}

\usepackage[pagebackref,breaklinks,colorlinks]{hyperref}

\usepackage{color, colortbl}
\definecolor{LightGray}{rgb}{0.92,0.92,0.92}
\definecolor{Gray1}{rgb}{0.95,0.95,0.95}
\definecolor{Gray2}{rgb}{0.9,0.9,0.9}
\definecolor{darkblue}{rgb}{0,0.592,0.655}
\definecolor{darkyellow}{rgb}{0.655,0.608,0}
\definecolor{redhl}{HTML}{ea9999}
\definecolor{greenhl}{HTML}{d9ead3}
\definecolor{bluehl}{HTML}{c9daf8}
\definecolor{yellowhl}{HTML}{fff2cc}

\usepackage{times,graphicx,tabulary,multirow,xspace,xcolor}

\usepackage{pifont}
\usepackage{changepage}

\newcommand{\correctanswer}{{\color[rgb]{0.01,0.38,0.75}\textbf{Blue}} highlights the correct answers.}
\newcommand{\bluehighlight}{{\color[rgb]{0.01,0.38,0.75}\textbf{Blue}}}
\newcommand{\redhighlight}{{\color[rgb]{0.75,0.0,0.0}\textbf{Red}}}

\newcommand{\eat}[1]{}
\newcommand{\modelname}{\textsc{MM-Vid}\xspace}
\newcommand{\gpttextonly}{GPT-4\xspace}
\newcommand{\gptmodelname}{GPT-4V\xspace}
\newcommand{\gptmodelnamefull}{GPT-4V(ision)\xspace}

\makeatletter
\DeclareRobustCommand\onedot{\futurelet\@let@token\@onedot}
\def\@onedot{\ifx\@let@token.\else.\null\fi\xspace}

\newcommand{\Paragraph}[1]{\vspace{0mm} \noindent \textbf{#1} \hspace{0mm}}

\usepackage[capitalize]{cleveref}
\crefname{section}{Sec.}{Secs.}
\Crefname{section}{Section}{Sections}
\Crefname{table}{Table}{Tables}
\crefname{table}{Tab.}{Tabs.}

\def\cvprPaperID{****} 
\def\confName{****}
\def\confYear{****}

\begin{document}

\title{
\modelname~\includegraphics[height=15pt]{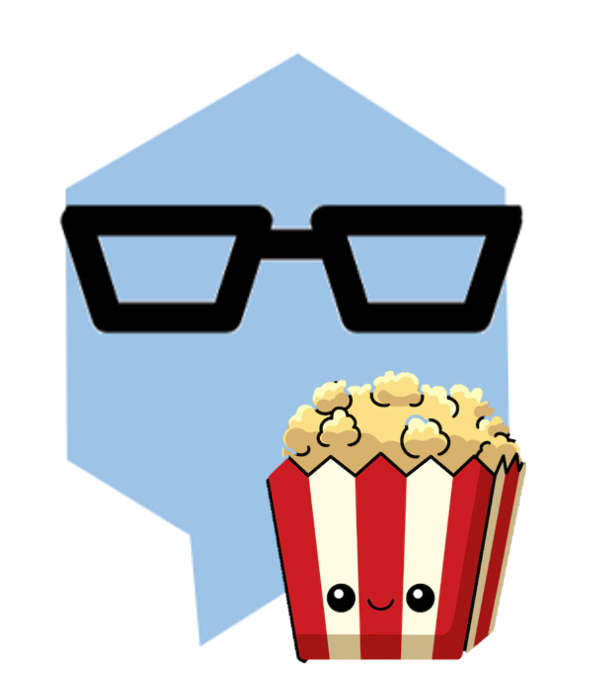}: Advancing Video Understanding with GPT-4V(ision)}

\author{Kevin Lin$^{*}$, Faisal Ahmed$^{*}$, Linjie Li$^{*}$, Chung-Ching Lin$^{*}$, Ehsan Azarnasab, Zhengyuan Yang, \\Jianfeng Wang, Lin Liang, Zicheng Liu, Yumao Lu, Ce Liu, Lijuan Wang$^{*\spadesuit}$\\Microsoft Azure AI\\
\small{
$^*$~Core Contribution \;
$^{\spadesuit}$~Project Lead \;
} \\ \url{https://multimodal-vid.github.io/}\\} 

\twocolumn[{
\renewcommand\twocolumn[1][]{#1}
\maketitle
\centering

\vspace{1em}
\captionsetup{type=figure}
\includegraphics[width=1.\textwidth]{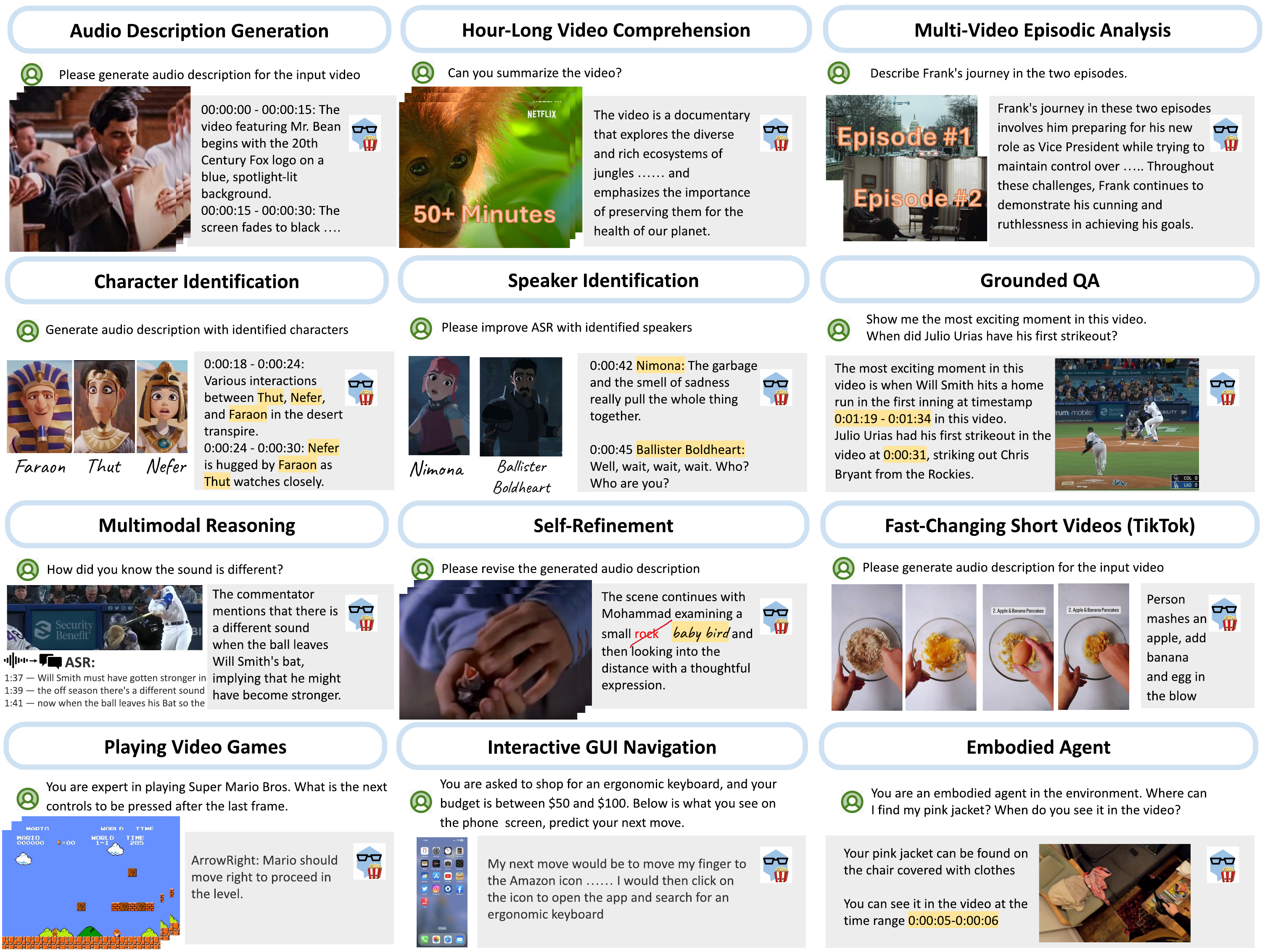}
\vspace{-1em}
\captionof{figure}{\modelname~allocates specialized vision, audio, speech experts with \gptmodelnamefull to address challenging video understanding tasks. For example, the system could associate information from multiple uploaded episodes and reason the storyline of the queried characters (``Multi-Video Episodic Analysis''). We highlight key information here and postpone full \modelname~responses to Figures~\ref{fig:vid-long-baseball-unfold-1}-\ref{fig:vid-gui-5}. Demo videos are available \href{https://multimodal-vid.github.io/\#video-demos}{at this link}.
\label{fig:teaser}
\vspace{3em}
}

}
]

\begin{abstract}
\vspace{-2mm}
We present \modelname, an integrated system that harnesses the capabilities of \gptmodelname\footnote{In this work, we explore \gptmodelnamefull~with the vision capability and refers to the model as ``\gptmodelname,'' following the OpenAI reports~\cite{gpt4v,gpt4}. We refer to the text-only version of the model as ``GPT-4''~\cite{gpt4}.}, combined with specialized tools in vision, audio, and speech, to facilitate advanced video understanding. \modelname is designed to address the challenges posed by long-form videos and intricate tasks such as reasoning within hour-long content and grasping storylines spanning multiple episodes.
\modelname uses a video-to-script generation with \gptmodelname to transcribe multimodal elements into a long textual script. The generated script details character movements, actions, expressions, and dialogues, paving the way for large language models (LLMs) to achieve video understanding. This enables advanced capabilities, including audio description, character identification, and multimodal high-level comprehension. Experimental results demonstrate the effectiveness of \modelname in handling distinct video genres with various video lengths. Additionally, we showcase its potential when applied to interactive environments, such as video games and graphic user interfaces.

\vspace{0pt}
\end{abstract}
\section{Introduction}

People around the world create numerous videos on a daily basis~\cite{pires2015youtube,miech19howto100m,carreira2019short,karpathy2014large}, including user-generated live streams, video-game live streams, short clips, movies, sports broadcasts, advertising, and more. Videos serve as a versatile medium for conveying information and content through various modalities~\cite{yuan2021florence,baltruvsaitis2018multimodal,snoek2005multimodal,yang2023code,zellers2021merlot,xu2023multimodal,shukor2023unified}, such as text, visuals, and audio. Developing methods that can learn from diverse modalities will enable us to design cognitive machines with enhanced capabilities for analyzing uncurated real-world videos, extending beyond the confines of hand-curated datasets. However, this rich representation introduces many challenges for the study of video understanding, particularly when dealing with extended-duration videos~\cite{wu2021towards,song2023moviechat}.  

Understanding long videos, especially those spanning over an hour, is a complex task that demands advanced methods capable of analyzing sequences of images and audio across multiple episodes. This challenge is compounded by the need to extract information from various sources, such as distinguishing speakers~\cite{nagrani2017voxceleb,snyder2019speaker,chen2021continuous}, identifying characters~\cite{kukleva2020learning,nagrani2018benedict,mark2006hello}, and maintaining narrative coherence~\cite{han2023autoad,rohrbach2015dataset}. Additionally, answering questions based on video evidence~\cite{lei2018tvqa} requires a deep comprehension of the content, context, and subtitles. When it comes to live streaming and gaming videos~\cite{pires2015youtube,baker2022video,dota2}, there are challenges in processing dynamic environments in real-time, requiring semantic understanding, and the ability of long-term strategy planning~\cite{baker2022video,park2023generative,xu2023exploring,wang2023survey,zhang2023building}.

Recently, substantial advances have been made with large pre-trained video models~\cite{feichtenhofer2019slowfast,arnab2021vivit,bertasius2021space,Fan_2021_ICCV,liu2022video, wang2017spatiotemporal} and video-language models~\cite{Lei_2021_CVPR,li2020hero,li2021value,bain2021frozen,lin2022swinbert,fu2021violet,lin2022egocentric,wang2023all,fu2023empirical,li2023lavender}, which have demonstrated their reasoning capabilities for video content. However, these models are usually trained on short clips (\textit{e.g.,} 10-second videos in Kinetics~\cite{carreira2017quo} and VATEX~\cite{Wang_2019_ICCV}) or pre-defined action classes (\textit{e.g.,} 174 classes in Something-Something v1~\cite{goyal2017something}). Consequently, these models may fall short in providing a detailed comprehension of intricate videos in real world~\cite{wu2021towards,song2023moviechat}. To achieve a more comprehensive understanding of the videos we encounter in daily life, we need methods capable of addressing complex challenges. It involves not only identifying who are in the scene and what they do, but also pinpointing when and how they act, while recognizing subtle nuances and visual cues across different scenes. The aim of this work is to address these challenges and explore methods that can be applied directly to real-world video understanding. Our approach involves breaking down extended video content into coherent narratives and subsequently employing these generated stories for video analysis.

\begin{figure*}[t]
\centering
\includegraphics[width=.99\textwidth]{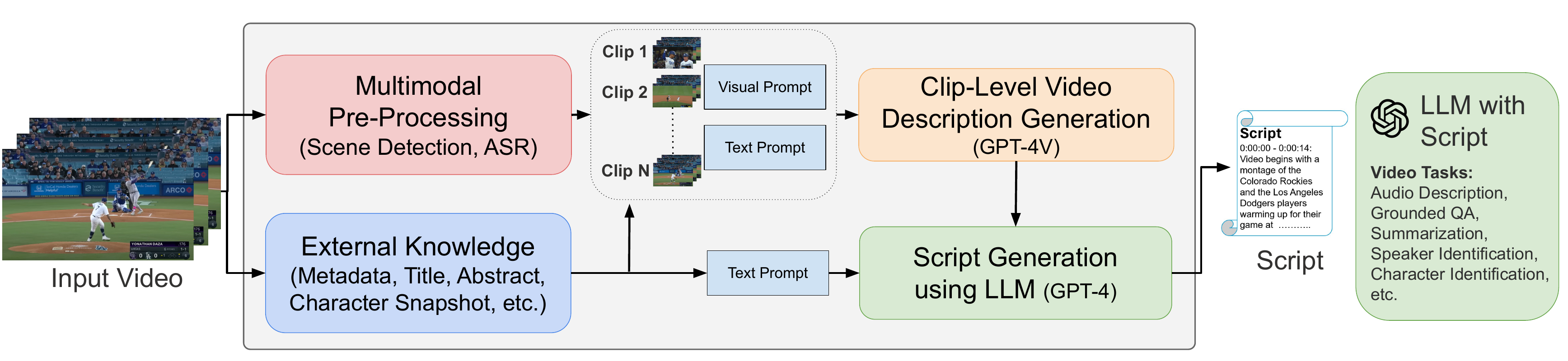}
\caption{Overview of \modelname. Our system takes a video file as input, and outputs a long textual script describing the video contents. \modelname consists of four modules: \textit{(i) Multimodal Pre-Processing}, \textit{(ii) External Knowledge Collection}, \textit{(iii) Clip-Level Video Description Generation}, and \textit{(iv) Script Generation}. 
}
\label{fig:arch}
\end{figure*}

Recent advances in Large Multimodal Models (LMMs)~\cite{alayrac2022flamingo,gpt4,gpt4v,yang2023dawn,bingchat,chowdhery2022palm,anil2023palm}, such as \gptmodelnamefull~\cite{gpt4v}, have demonstrated significant breakthroughs in processing both input images and text for multimodal understanding. This has sparked interest in applying LMMs to the video domain. In this work, we present \modelname, a system that integrates specialized tools with \gptmodelname for video understanding. Given an input video, \modelname performs multimodal pre-processing, including scene detection and automatic speech recognition (ASR), to collect important information in the video. The input video is then split into multiple clips according to the scene detection algorithm. Then, we employ \gptmodelname, which takes the clip-level video frames as input and generates a detailed description for each video clip. Finally, \gpttextonly is adopted to generate a coherent script for the full video, conditioning on the clip-level video descriptions, ASR, and video metadata if available. As shown in Figure~\ref{fig:teaser}, the generated script allows \modelname to perform a diverse set of video tasks. 

Experimental results demonstrate the effectiveness of \modelname in different challenging scenarios. \modelname is able to comprehend hour-long videos through multiple modalities, and localize specific events with correct timestamps.
\modelname also demonstrates intriguing results in an interactive environment, such as predicting the possible next steps when playing a video game~\cite{pygame} or interacting with a graphical user interface (GUI)~\cite{yang2023dawn}.

\section{Related Work}
\Paragraph{Conventional Video Understanding Methods.} Early work in computer vision centered on building video foundation models~\cite{feichtenhofer2019slowfast,arnab2021vivit,bertasius2021space,Fan_2021_ICCV,liu2022video, wang2017spatiotemporal}. These models, with different neural network architecture designs and training methods, have achieved great breakthrough at analyzing short video clips~\cite{carreira2017quo,carreira2019short,soomro2012ucf101,kuehne2011hmdb}, typically lasting less than 30 seconds. However, these models are typically pre-trained with vision modality only, and thus may require specific adjustment or fine-tuning for multimodal downstream tasks.

\Paragraph{Video-Language Models.} Recent studies~\cite{Lei_2021_CVPR,li2020hero,li2021value,bain2021frozen,lin2022swinbert,fu2021violet,lin2022egocentric,wang2023all,fu2023empirical,li2023lavender}  have made remarkable improvements in multimodal representation learning for video-and-language understanding. These advancements have been particularly evident in popular downstream tasks such as video question answering~\cite{lei2018tvqa}, text-video retrieval~\cite{lei2020tvr,xu2016msr} and video captioning~\cite{Wang_2019_ICCV}. Building on this momentum, researchers typically embark on a pretrain-finetune paradigm: initially pre-training a video-language foundation model on large-scale video-text pairs, followed by a fine-tuning process on specific downstream datasets. However, these methods are usually trained on short video clips, often restricted to durations of around 10 seconds, posing potential challenges in comprehending longer video sequences.

\Paragraph{Visual Instruction Tuning.} Inspired by the breakthrough of Large Language Models (LLMs)~\cite{gpt4,chowdhery2022palm,touvron2023llama,vicuna2023,zhang2205opt}, recent studies~\cite{zhang2023video,maaz2023video,luo2023valley,song2023moviechat,2023videochat} suggest using a frozen LLM combined with an image encoder and a few learnable modules for video understanding tasks. Specifically, researchers propose the visual instruction tuning~\cite{liu2023llava,maaz2023video,2023videochat}, which aims to fine-tune the learnable modules and thus enable LLMs to generate textual descriptions for the video content. 
While promising performance is presented, these models may fall short when it comes to handling videos with extended duration. Our work aims to fill this gap, exploring methods that can be directly applied to the understanding of long videos in real-world situations.

\Paragraph{Prompting LLMs for Video Understanding.} Recently, researchers~\cite{vlog2023,wang2023chatvideo,xie2022visual,li2023multimodal} explore the LangChain system paradigm~\cite{langchain}, which aims to integrate expert tools with existing LLMs to create new functionalities. For example, VLog~\cite{vlog2023} uses BLIP2~\cite{li2023blip} and GRIT~\cite{wu2022grit} as dense image captioners, Whisper~\cite{radford2023robust} as ASR translator, and ChatGPT as a reasoner. By transcribing a given video to textual descriptions (\textit{e.g.,} document), it enables ChatGPT for video question-answering tasks. Inspired by the efficacy of these tool-using approaches~\cite{langchain,yang2023mm,wang2023chatvideo}, we explore integration with \gptmodelname for video understanding.

\section{Preliminary Study with \gptmodelnamefull}

Recent studies~\cite{gpt4,gpt4v,gpt4vblog,yang2023dawn} show that \gptmodelname can accept a range of inputs, such as textual descriptions, questions, or even visual cues like images or short video clips. \gptmodelname's inherent ability to comprehend visual inputs and generate contextually relevant text opens the door for a wide range of applications. By introducing a sequence of frames as input, \gptmodelname can grasp temporal relationships and interactions, aiding in the identification and interpretation of dynamic visual content.

\section{\textbf{\modelname}}

Figure~\ref{fig:arch} shows the overview of our system pipeline. \modelname takes the video file as input, and outputs a script describing the video contents. The generated script enables LLMs to achieve various video understanding capabilities.
\modelname consists of four modules: \textit{(i) Multimodal Pre-Processing}, \textit{(ii) External Knowledge Collection}, \textit{(iii) Clip-Level Video Description Generation}, and \textit{(iv) Script Generation}. We describe each module in detail below.

\Paragraph{Multimodal Pre-Processing.} Starting with an input video file, our process begins by using the established ASR tool to extract transcriptions from the video. Following this, we divide the video into several short video clips. This process involves uniform sampling of video frames, with each clip consisting of 10 frames. To enhance the overall quality of frame sampling, we use established scene detection tools like PySceneDetect~\cite{pyscenedet} to help identify crucial scene boundaries.

\Paragraph{External Knowledge Collection.} We incorporate external knowledge into our input prompts to \gptmodelname. This involves gathering available information, such as metadata, title, abstract, and face photos of characters within the video. In our experiments, the metadata, title, and abstract are gathered from YouTube.

\Paragraph{Clip-Level Video Description Generation.} During our multimodal pre-processing, the input video is segmented into multiple clips. For each clip, which typically consists of 10 frames, we employ \gptmodelname to generate video descriptions. By feeding the video frames along with the associated text prompt into the model, \gptmodelname utilizes the input to generate detailed descriptions that capture the visual elements, actions, and events depicted in those frames.

In addition, we explore the use of visual prompting, where the character's face photos are presented alongside the character's name in the input to \gptmodelname. Our empirical results suggest that visual prompting is helpful to enhance the quality of video descriptions, particularly for more accurate character identification. These findings align with the insights from~\cite{yang2023dawn}.

\Paragraph{Script Generation using LLM.} After generating the descriptions for each video clip, we use \gpttextonly to integrate these clip-level descriptions into a coherent script. This script serves as a comprehensive description of the entire video, and is used by \gpttextonly for a diverse set of video understanding tasks.

\begin{figure}[t]
\centering
\includegraphics[width=.99\columnwidth]{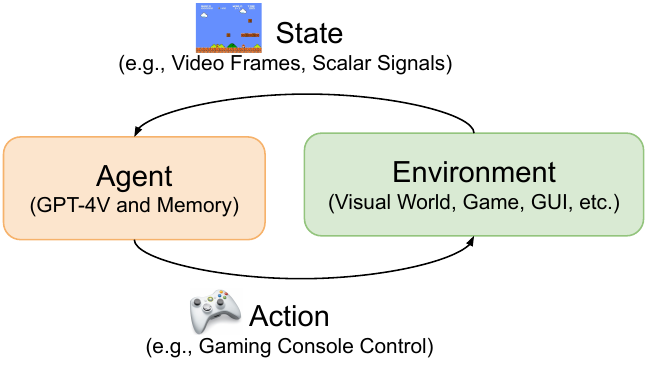}
\caption{\modelname for streaming inputs. \modelname can serve as an agent in an interactive environment, continually receiving and processing the streaming video frames.  
}
\label{fig:arch2}
\end{figure}

\section{\modelname for Streaming Inputs}
Figure~\ref{fig:arch2} shows the diagram of \modelname when applied to the context of streaming inputs. Our system operates as an agent within a dynamic environment where streaming video frames serve as the primary input. In this context, the agent continually receives streaming video frames as states, representing the ongoing visual information unfolding in the environment. These states are then processed by \gptmodelname to make informed decisions and generate responses.

By continually analyzing the streaming video frames, \modelname plays a crucial role in transforming raw visual data into meaningful insights, making it valuable for applications such as video game play, the embodied agent, and GUI navigation.

\begin{figure*}[h]
\centering
\vspace{-40pt}
\includegraphics[width=0.92\textwidth]{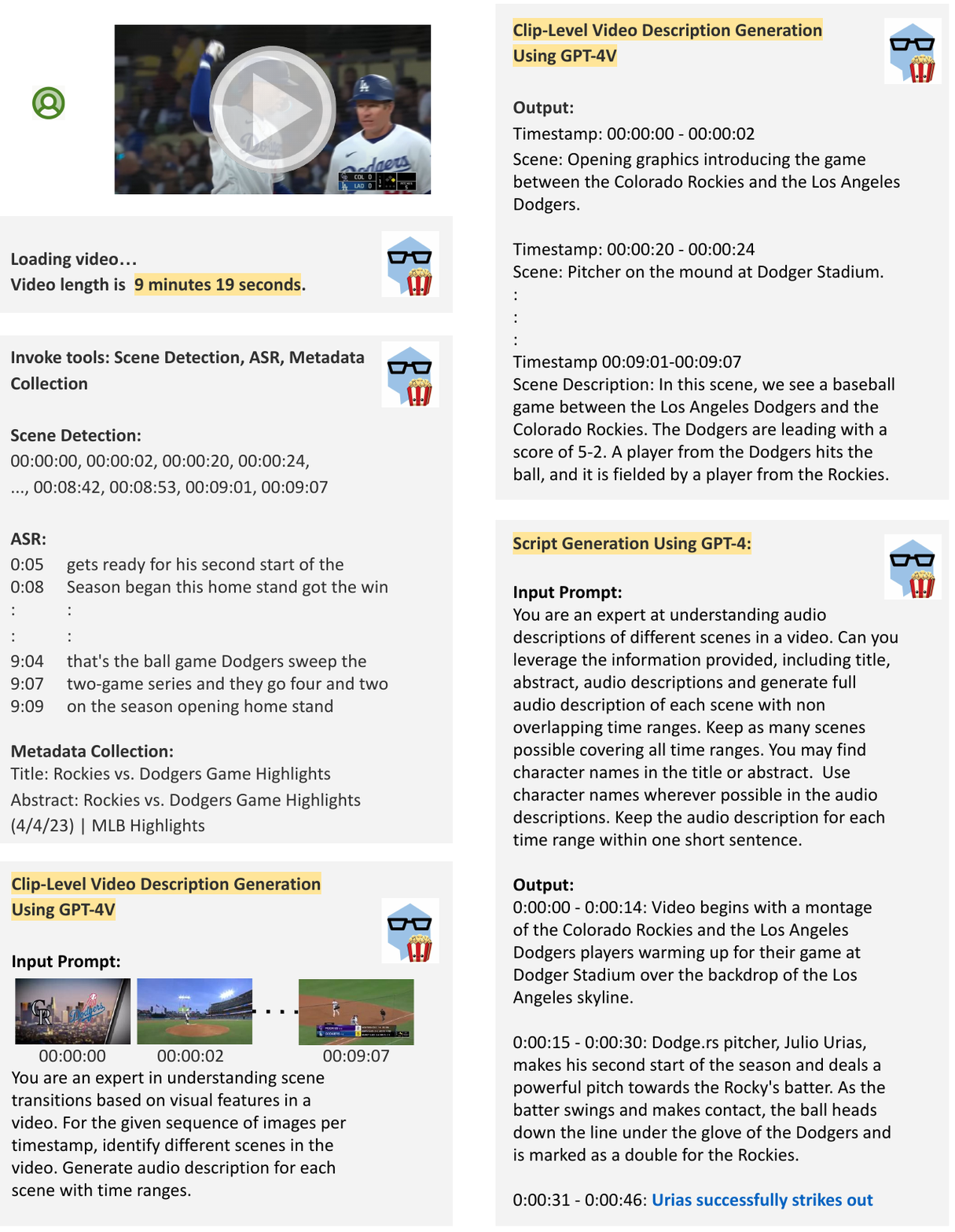}
\caption{An example of \modelname's execution flow. Given a baseball video, \modelname provides an estimated video length, and then invokes scene detection and ASR tools, and collects external knowledge. Then, we generate clip-level video descriptions by using \gptmodelname. \gptmodelname takes the video frames and the text prompt as input, and outputs the video descriptions. Finally, given the clip-level descriptions, video metadata and ASR, we use \gpttextonly to generate a coherent script for the input video. Figures \ref{fig:vid-long-baseball-unfold-2}-\ref{fig:vid-long-baseball-unfold-4} show continued outputs. The original video is available at \url{https://www.youtube.com/watch?v=-pNe0p4H8ec}
}
\label{fig:vid-long-baseball-unfold-1}
\end{figure*}

\section{Experiments}
\subsection{Experiment Setup}

We implement \modelname based on \textsc{MM-ReAct}~\cite{yang2023mm} codebase. We use the Automatic Speech Recognition (ASR) tool publicly available via the Azure Cognitive Services APIs~\cite{azureapi}, and utilize PySceneDetect~\cite{pyscenedet} for scene detection. 

\subsection{\textbf{\modelname}~Capabilities}

\begin{figure*}[th]
\centering
\vspace{-40pt}
\includegraphics[width=1.0\textwidth]{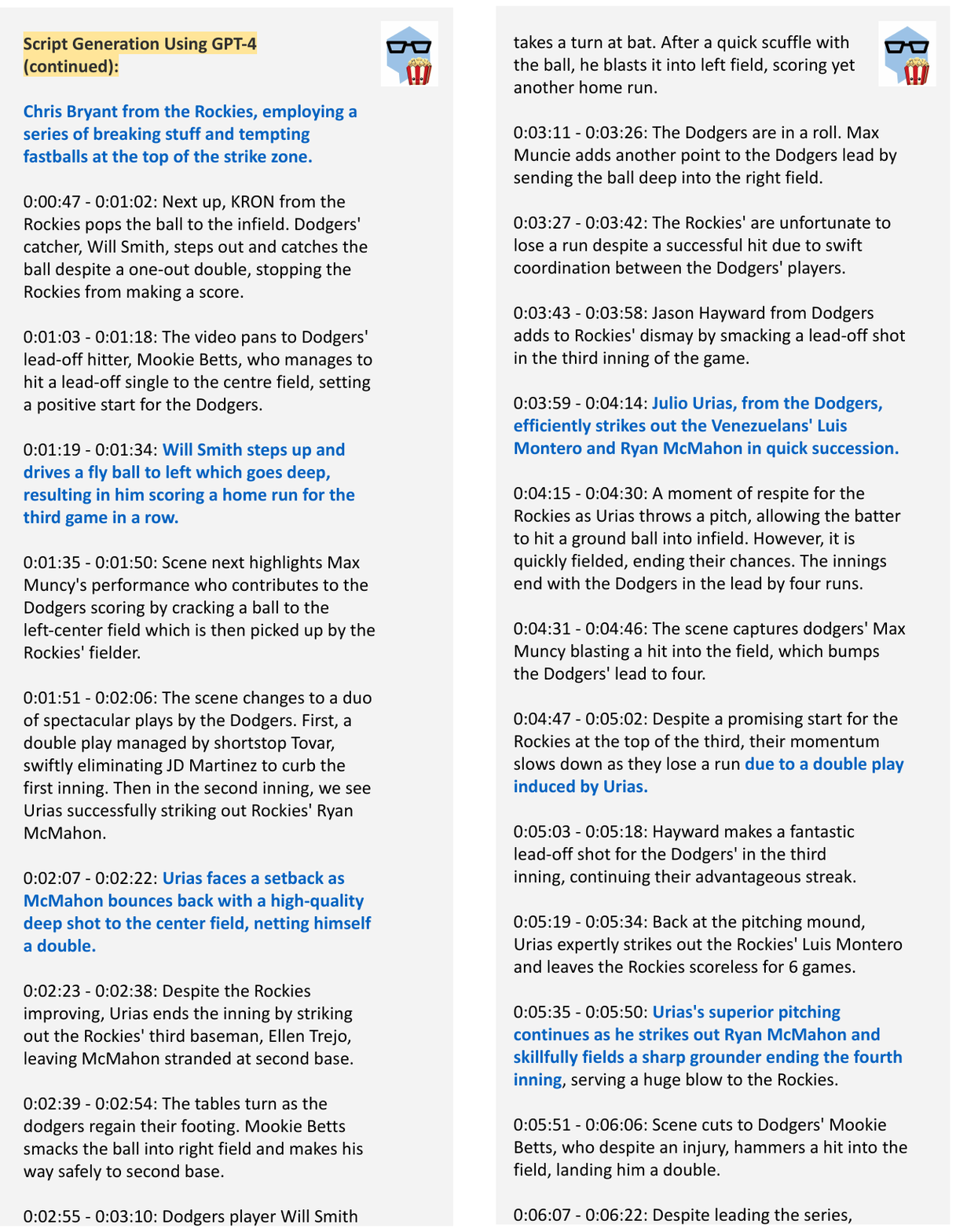}
\vspace{-1ex}
\caption{An example of \modelname's execution flow (continued). We present the full script generated by \modelname. The original video is available at \url{https://www.youtube.com/watch?v=-pNe0p4H8ec}
}
\label{fig:vid-long-baseball-unfold-2}
\end{figure*}

\begin{figure*}[th]
\centering
\vspace{-40pt}
\includegraphics[width=1.0\textwidth]{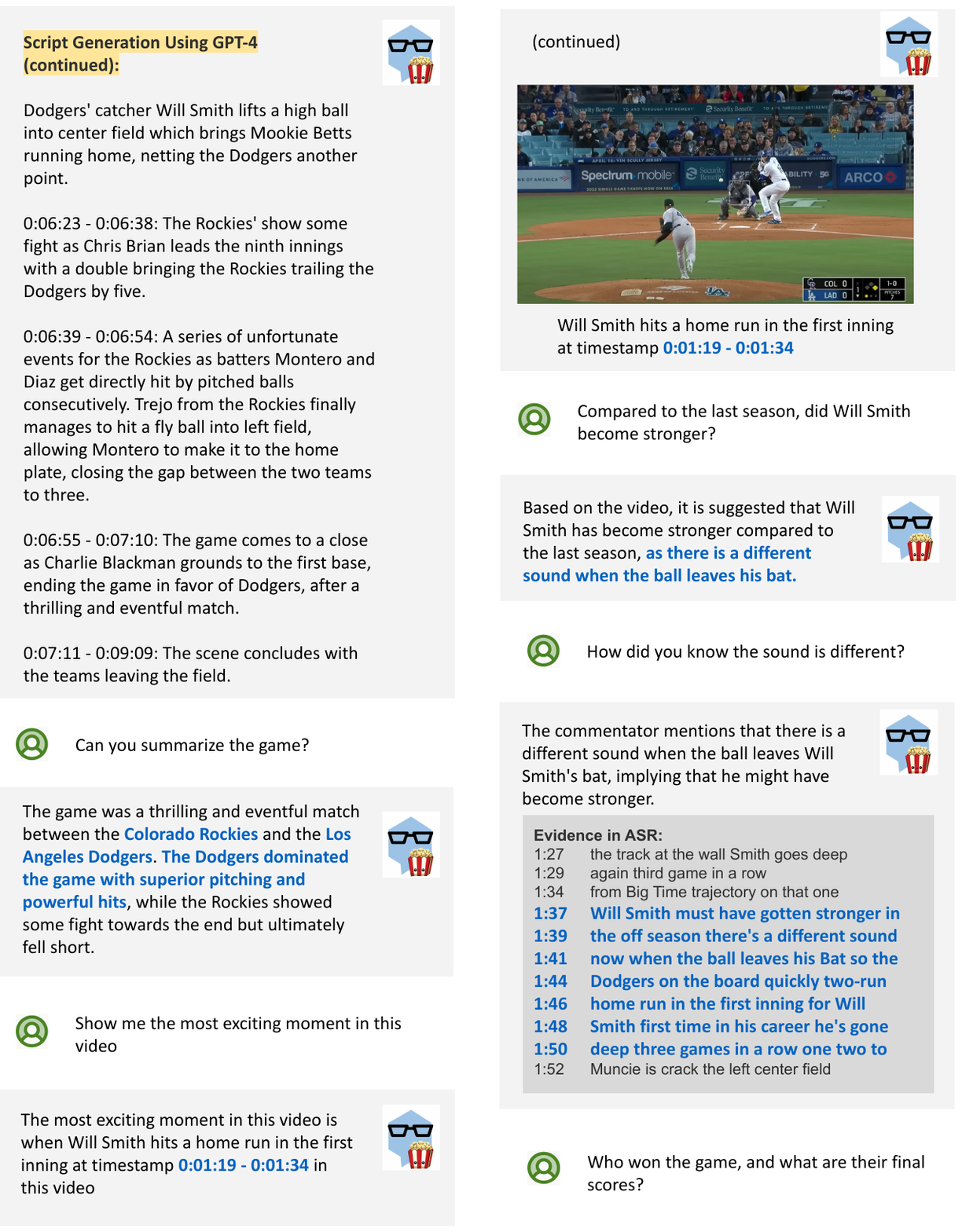}
\vspace{-1ex}
\caption{An example of \modelname's execution flow (continued). Given the generated script as reference, \modelname achieves various video understanding capabilities, including \textbf{grounded QA} and \textbf{multimodal reasoning}. The original video is available at \url{https://www.youtube.com/watch?v=-pNe0p4H8ec}
}
\label{fig:vid-long-baseball-unfold-3}
\end{figure*}

\begin{figure*}[th]
\centering
\vspace{-40pt}
\includegraphics[width=1.0\textwidth]{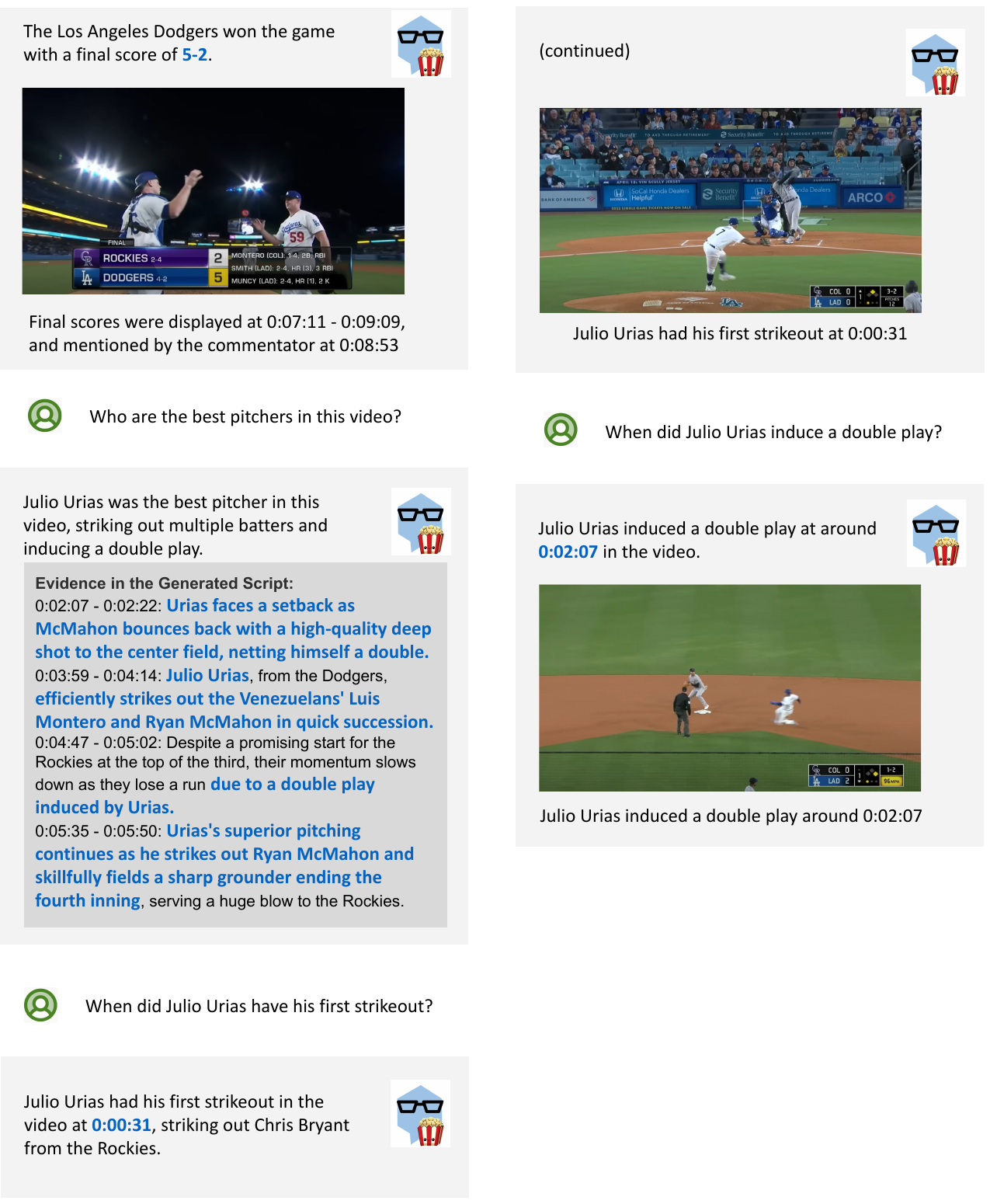}
\vspace{-1ex}
\caption{An example of \modelname's execution flow (continued). Given the generated script as reference, \modelname achieves various video understanding capabilities, including \textbf{grounded QA} and \textbf{multimodal reasoning}. The original video is available at \url{https://www.youtube.com/watch?v=-pNe0p4H8ec}
}
\label{fig:vid-long-baseball-unfold-4}
\end{figure*}

Figures~\ref{fig:vid-long-baseball-unfold-1}-\ref{fig:vid-long-baseball-unfold-4} provide illustrative examples of \modelname's complete execution flow. When a user uploads a video file, \modelname initiates the process by first assessing the estimated video length. Subsequently, it performs multimodal pre-processing by invoking expert tools, including scene detection and ASR. Additionally, \modelname collects external knowledge, encompassing video metadata such as title and abstract.

Following this preliminary stage, \modelname proceeds to generate clip-level video descriptions for each segment of the video. Finally, it invokes \gpttextonly, integrating these clip-level descriptions into a coherent script. Once the script is generated, it empowers LLMs to provide a summarized understanding of the video content. That equips the system to address users' questions with grounded answers. We discuss \modelname's distinct capabilities as below.

\Paragraph{Grounded Question-Answer (QA).} The generation of a comprehensive script empowers our system with the capability of grounded QA. As shown in Figure~\ref{fig:vid-long-baseball-unfold-3}, let us consider a scenario where a user poses the question, ``Show me the most exciting moment in this video.'' In response, \modelname displays a highlight, specifically featuring a home run, and provides the corresponding timestamp. When a user asks ``Who are the best pitchers in this video?'' \modelname addresses the question by referring to relevant evidence in the generated script. This grounding capability owes its success to the extensive and detailed script generation process, which documents essential timestamps and significant events within the video, enabling accurate and contextually grounded responses to user inquiries.

\Paragraph{Multimodal Reasoning.} \modelname considers multimodal inputs, including video frames, speech transcriptions, and external knowledge if available. In Figure~\ref{fig:vid-long-baseball-unfold-3}, when a user inquires, ``How did you know the sound is different?'' \modelname explains that this information was derived from the commentator's remarks during the game. The examples illustrate \modelname's multimodal reasoning capabilities, where it integrates both visual and auditory cues to provide contextually accurate responses to user queries.
 
\begin{figure*}[th]
\centering
\vspace{-40pt}
\includegraphics[width=1.0\textwidth]{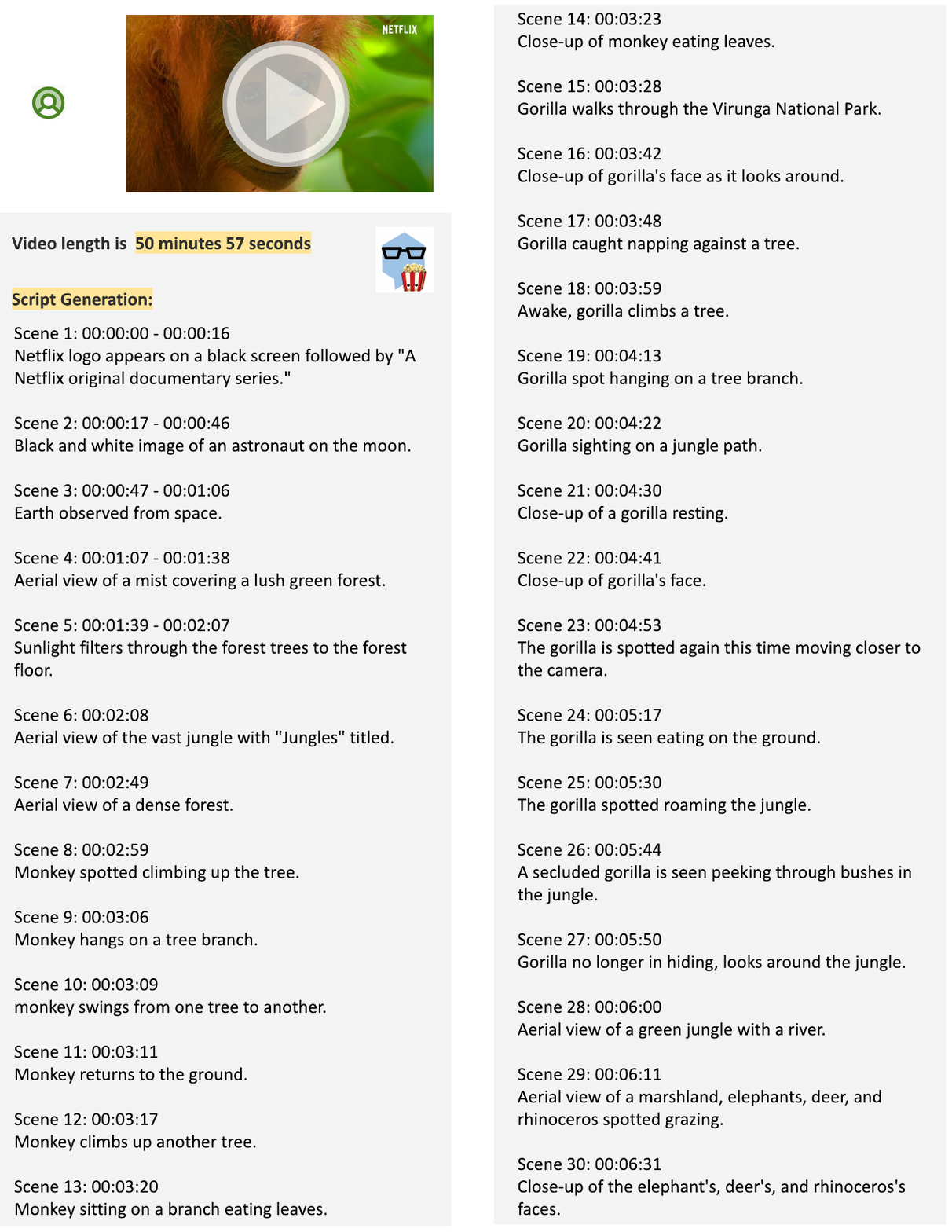}
\vspace{-1ex}
\caption{Case studies of \modelname’s capabilities and application scenarios: \textbf{hour-long video comprehension}.  Figures \ref{fig:vid-long-nature-2}-\ref{fig:vid-long-nature-4} show continued outputs. The original video is available at \url{https://www.youtube.com/watch?v=um2Q9aUecy0}
}
\label{fig:vid-long-nature-1}
\end{figure*}

\begin{figure*}[th]
\centering
\vspace{-40pt}
\includegraphics[width=1.0\textwidth]{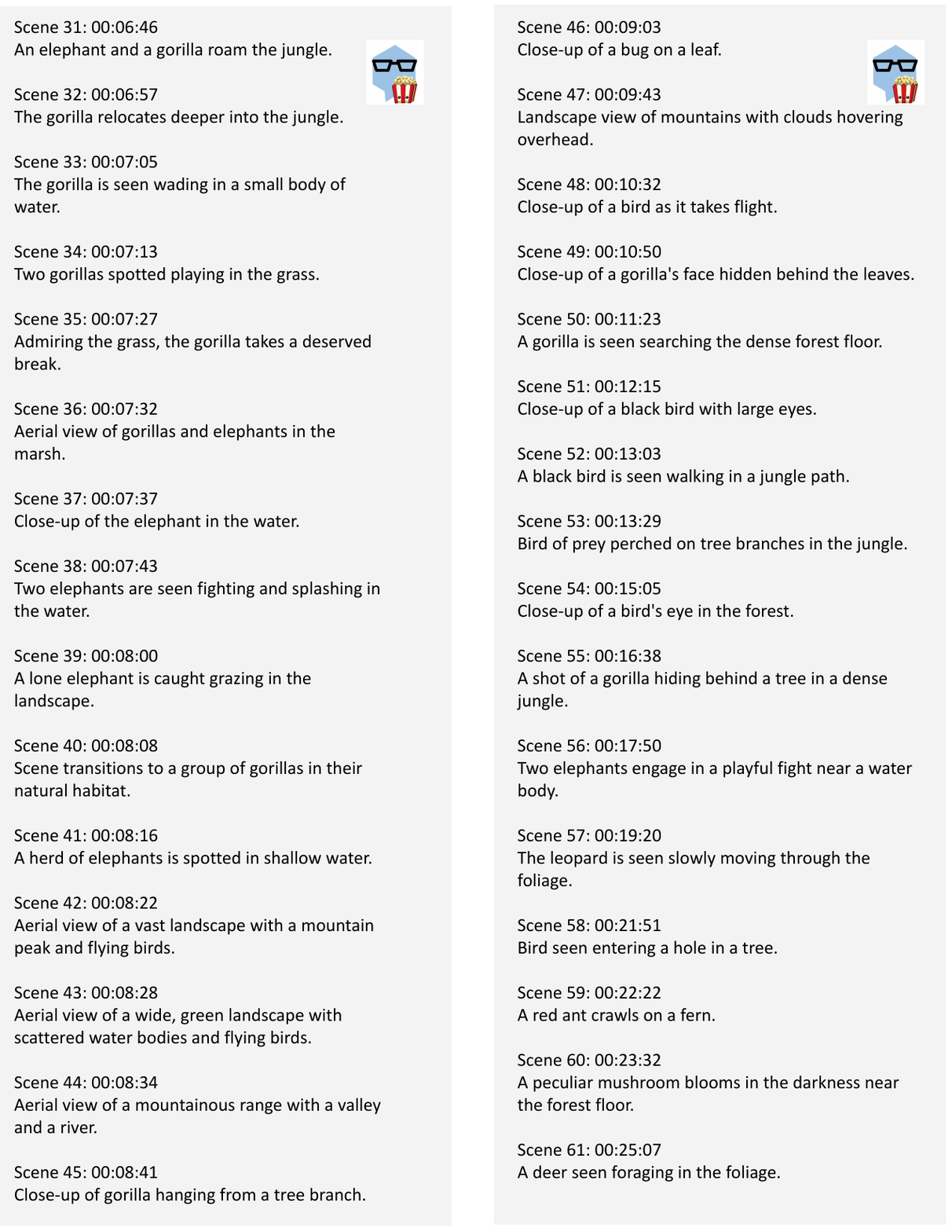}
\vspace{-1ex}
\caption{Case studies of \modelname’s capabilities and application scenarios: \textbf{hour-long video comprehension}; We present the full script generated by \modelname. The original video is available at \url{https://www.youtube.com/watch?v=um2Q9aUecy0}
}
\label{fig:vid-long-nature-2}
\end{figure*}

\begin{figure*}[th]
\centering
\vspace{-40pt}
\includegraphics[width=1.0\textwidth]{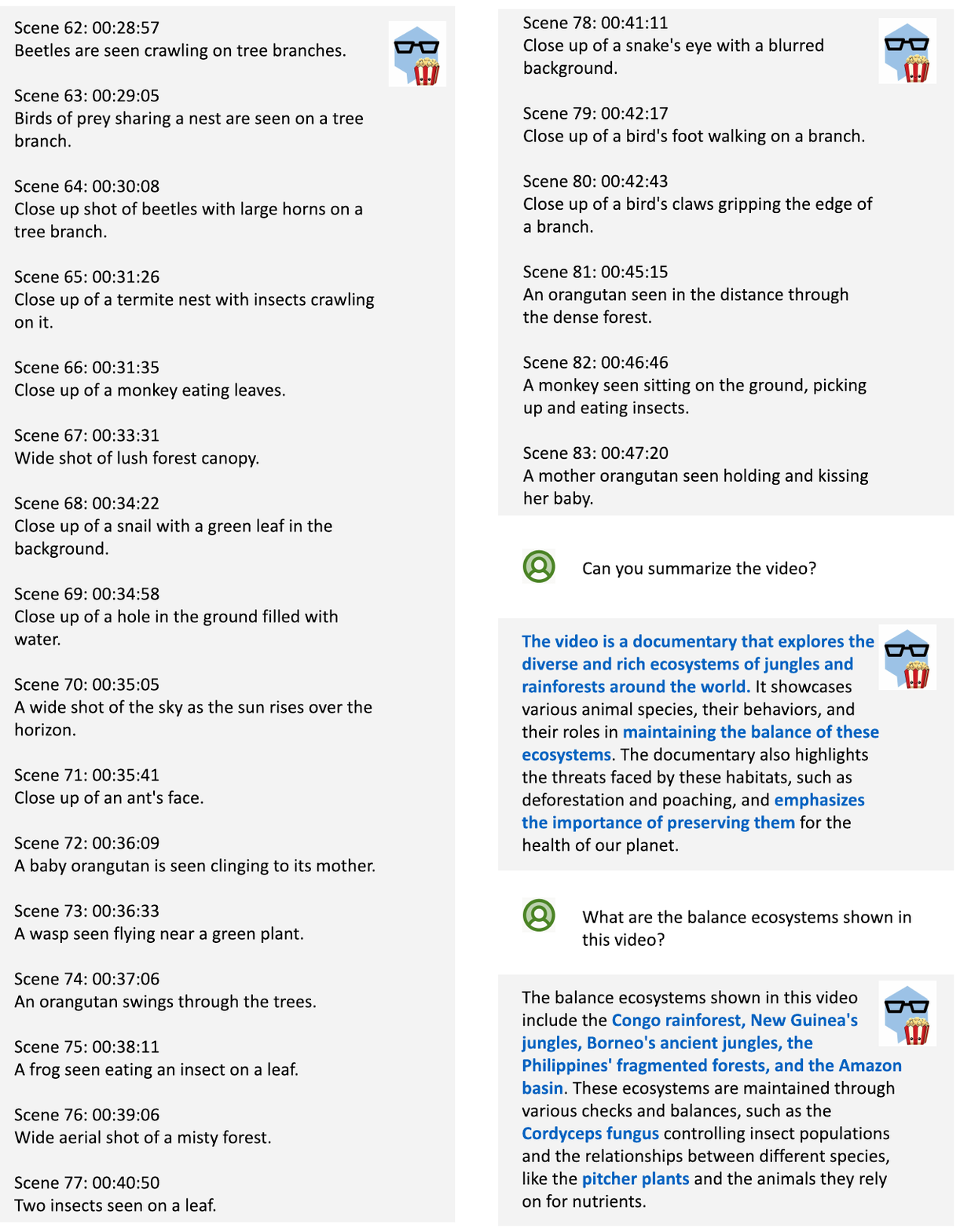}
\vspace{-1ex}
\caption{Case studies of \modelname’s capabilities and application scenarios: \textbf{hour-long video comprehension}; \modelname is able to summarize the video and address specific questions with regard to the video contents. \correctanswer \ \  The original video is available at \url{https://www.youtube.com/watch?v=um2Q9aUecy0}
}
\label{fig:vid-long-nature-3}
\end{figure*}

\begin{figure*}[th]
\centering
\vspace{-40pt}
\includegraphics[width=1.0\textwidth]{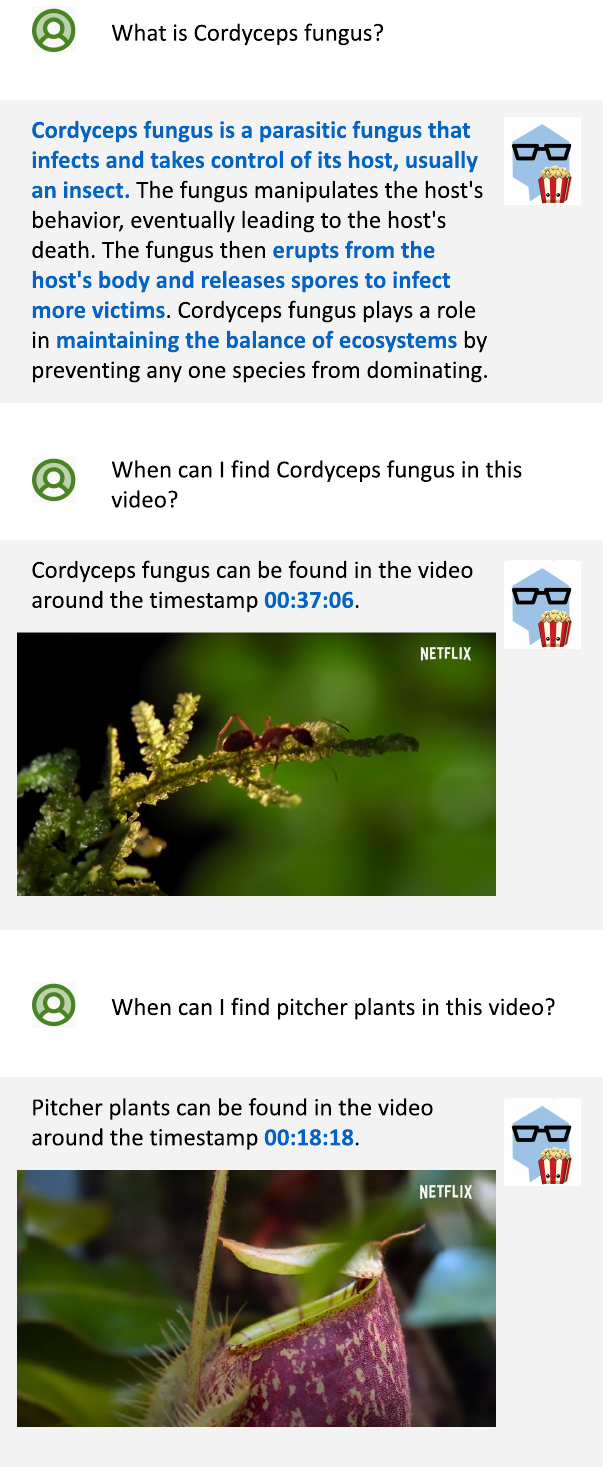}
\vspace{-1ex}
\caption{Case studies of \modelname’s capabilities and application scenarios: \textbf{hour-long video comprehension}; \modelname is able to localize the specific moments or topics in the lengthy documentary video. \correctanswer \ \ The original video is available at \url{https://www.youtube.com/watch?v=um2Q9aUecy0}
}
\label{fig:vid-long-nature-4}
\end{figure*}

\Paragraph{Hour-Long Video Comprehension.} Figures~\ref{fig:vid-long-nature-1}-\ref{fig:vid-long-nature-4} demonstrate \modelname's capabilities in processing lengthy videos. In this example, \modelname effectively analyzes a documentary video spanning approximately 50 minutes in duration. For simplicity, the intermediate outputs are omitted in the figures, and only the final generated script is presented. We observe that \modelname is able to generate a long script with the corresponding timestamps to represent the documentary video. By leveraging this generated script as contextual information, \modelname is equipped to perform a range of tasks, including summarizing the lengthy video, addressing specific queries raised within the video, and indexing pivotal moments.

\begin{figure*}[th]
\centering
\vspace{-40pt}
\includegraphics[width=1.0\textwidth]{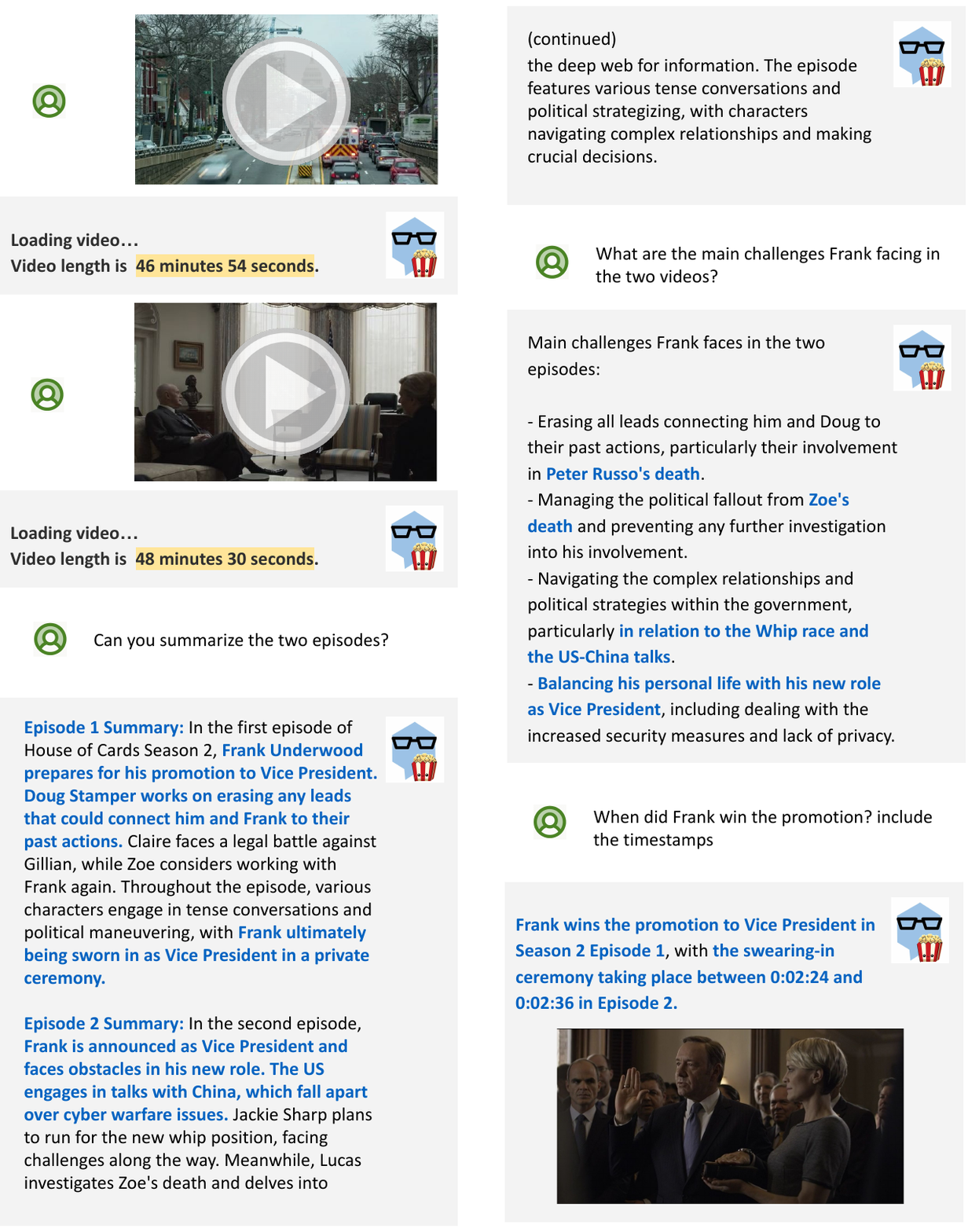}
\vspace{-1ex}
\caption{Case studies of \modelname’s capabilities and application scenarios: \textbf{multi-video episodic analysis}; Given multiple video files as inputs, \modelname can summarize the video series, perform cross-episodic reasoning, describe a specific character's journey across multiple videos. Figure \ref{fig:vid-long-eps-1-2} shows continued outputs. \correctanswer \ \ The two videos are from House of Cards, Season 2: episode 1 and episode 2.
}
\label{fig:vid-long-eps-1}
\end{figure*}

\begin{figure*}[th]
\centering
\vspace{-40pt}
\includegraphics[width=1.0\textwidth]{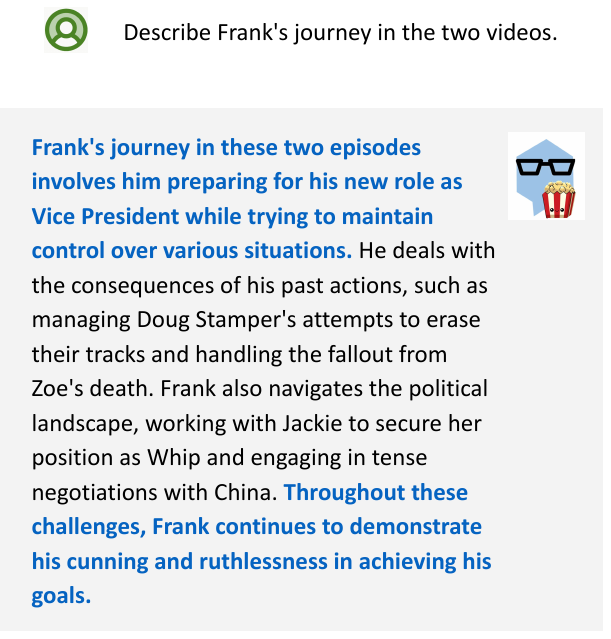}
\vspace{-1ex}
\caption{Case studies of \modelname’s capabilities and application scenarios: \textbf{multi-video episodic analysis}; Given multiple video files as inputs, \modelname can summarize the video series, perform cross-episodic reasoning, describe a specific character's journey across multiple videos. \correctanswer \ \ The two videos are from House of Cards, Season 2: episode 1 and episode 2.
}
\label{fig:vid-long-eps-1-2}
\end{figure*}

\begin{figure*}[th]
\centering
\vspace{-40pt}
\includegraphics[width=1.0\textwidth]{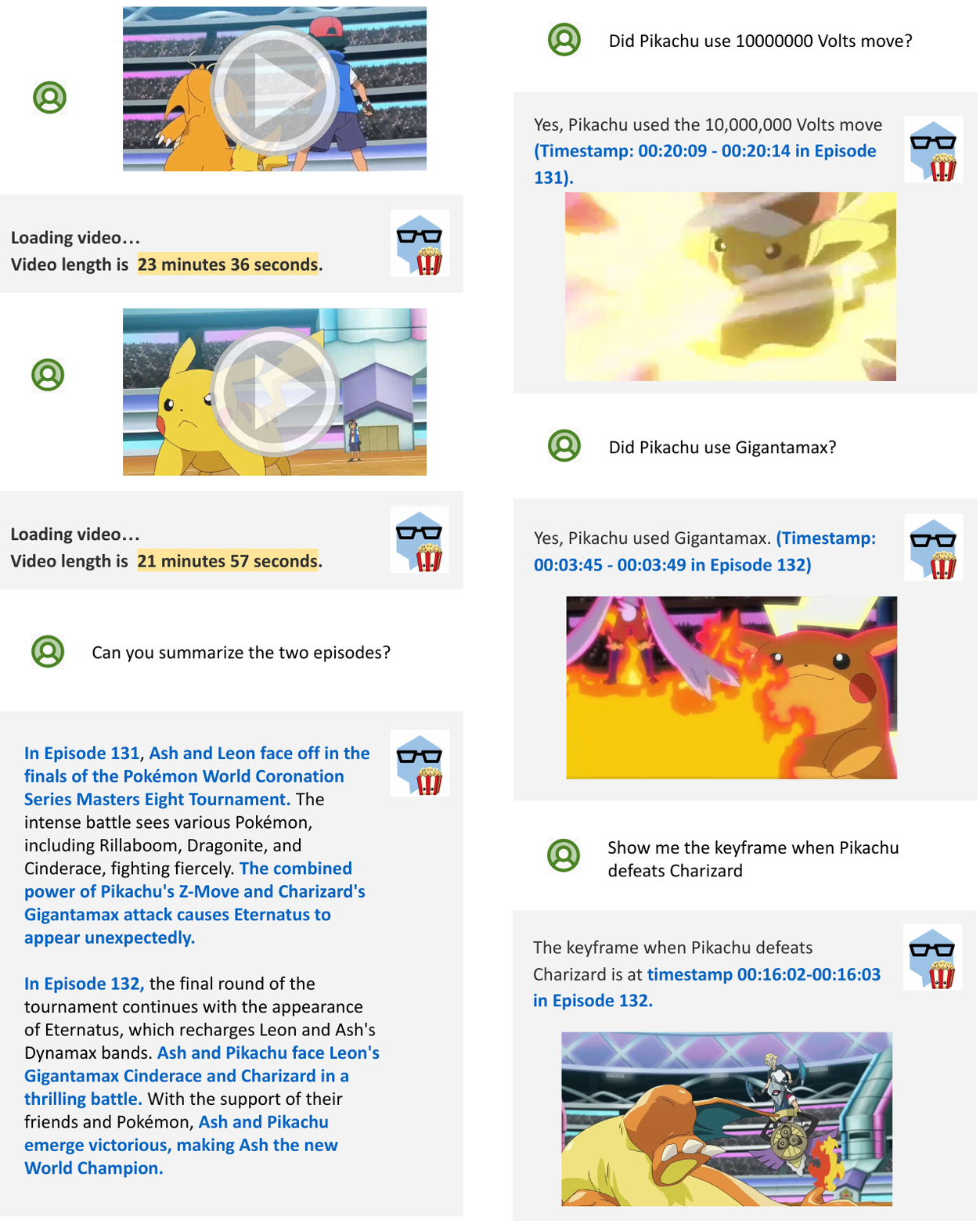}
\vspace{-1ex}
\caption{Case studies of \modelname’s capabilities and application scenarios: \textbf{multi-video episodic analysis}; Given multiple video files as inputs, \modelname can summarize the video series, and localize specific moments across multiple videos. \correctanswer \ \ The two videos are from Pokémon Journeys: episode 131 and episode 132.
}
\label{fig:vid-long-eps-2}
\end{figure*}

\Paragraph{Multi-Video Episodic Analysis.} \modelname's proficiency in handling extensive video content can be expanded to encompass multiple lengthy videos, as illustrated in Figures~\ref{fig:vid-long-eps-1}-\ref{fig:vid-long-eps-2}. In these examples, we upload multiple episodes to \modelname, showcasing its ability to perform a variety of complex tasks. \modelname exhibits the capability to summarize the video series, engage in cross-episode reasoning, provide detailed descriptions of character journeys across multiple episodes, and facilitate grounded QA interactions.

\begin{figure*}[th]
\centering
\vspace{-40pt}
\includegraphics[width=1.0\textwidth]{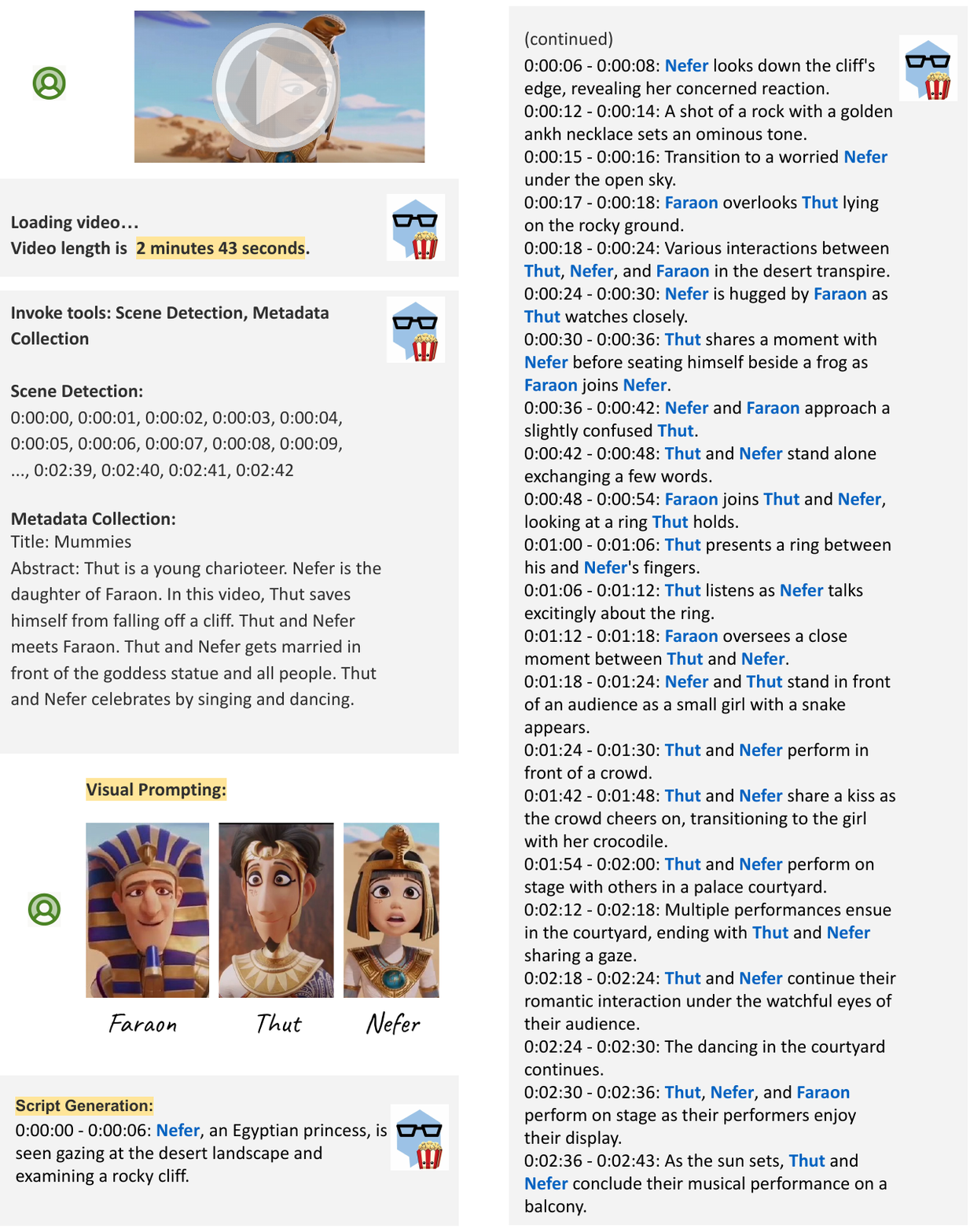}
\vspace{-1ex}
\caption{Case studies of \modelname’s capabilities and application scenarios: \textbf{character identification}; By providing the face photos of the characters and their names, \modelname can generate the script with the identified characters. \bluehighlight~highlights the correct character identification. The original video is available at \url{https://www.tokyvideo.com/video/mummies-online-free-film}, from 1:19:47 to 1:22:30.
}
\label{fig:vid-mummy-visual-prompt}
\end{figure*}

\Paragraph{Character Identification.} We found that incorporating visual prompts enhances the quality of script generation, particularly with regards to character identification. In Figure~\ref{fig:vid-mummy-visual-prompt}, we illustrate this by providing \modelname with additional inputs consisting of characters' face photos and their corresponding names. \modelname effectively utilizes these visual prompts to identify the characters depicted in the video, based on the provided face photos. As a result, the script generation process is notably improved, ensuring more accurate and contextually relevant descriptions of characters and their interactions within the video content. 

\begin{figure*}[th]
\centering
\vspace{-40pt}
\includegraphics[width=1.0\textwidth]{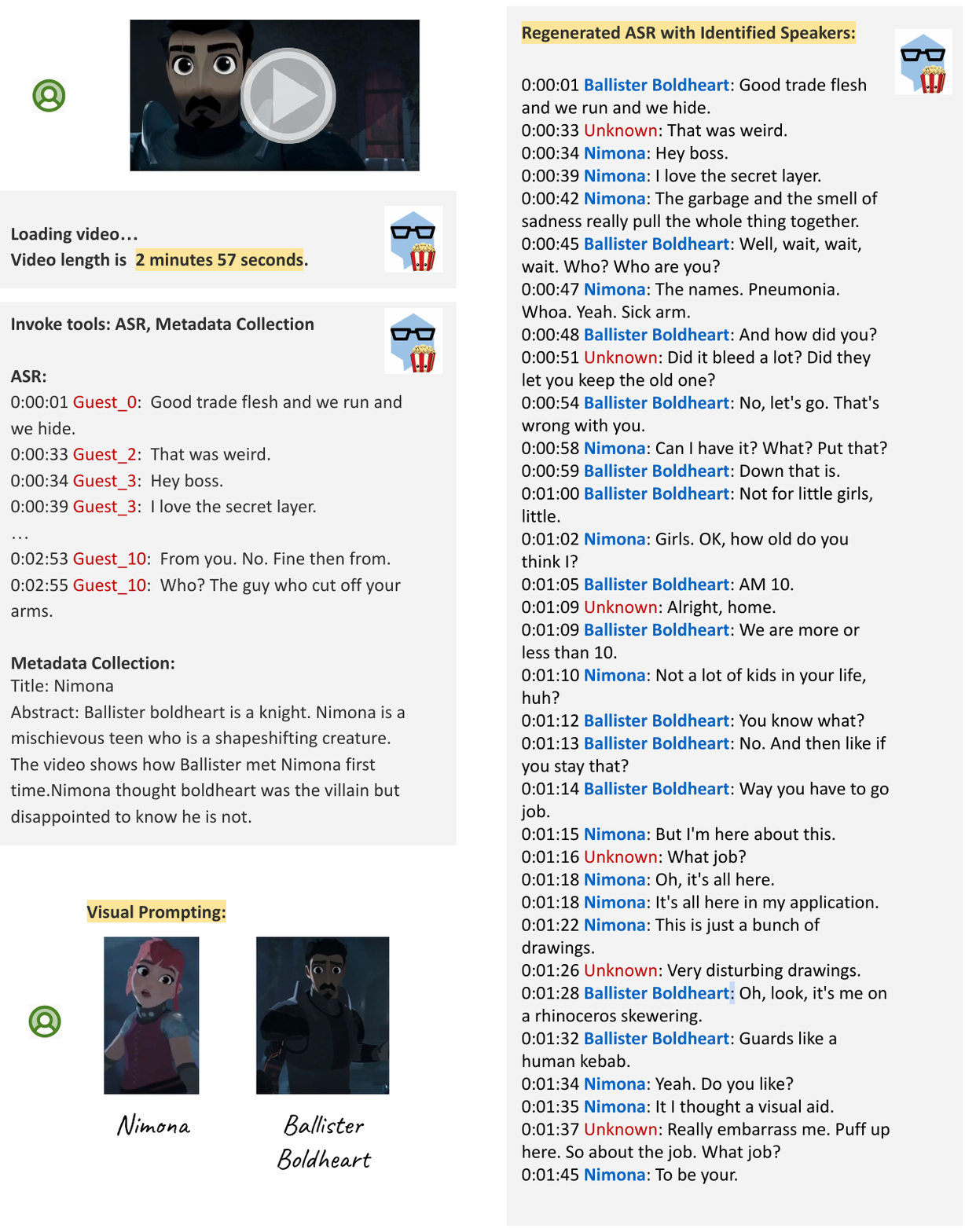}
\vspace{-1ex}
\caption{Case studies of \modelname’s capabilities and application scenarios: \textbf{speaker identification}; By leveraging visual prompting, \modelname can enhance ASR predictions with the speakers' identity. \bluehighlight~and~\redhighlight~highlight the correct and incorrect predictions, respectively. Figure \ref{fig:vid-speaker-id-2} shows continued outputs. The original video is available at \url{https://www.netflix.com/title/81444554}, from 9:52 to 12:52.
}
\label{fig:vid-speaker-id-1}
\end{figure*}

\begin{figure*}[th]
\centering
\vspace{-40pt}
\includegraphics[width=1.0\textwidth]{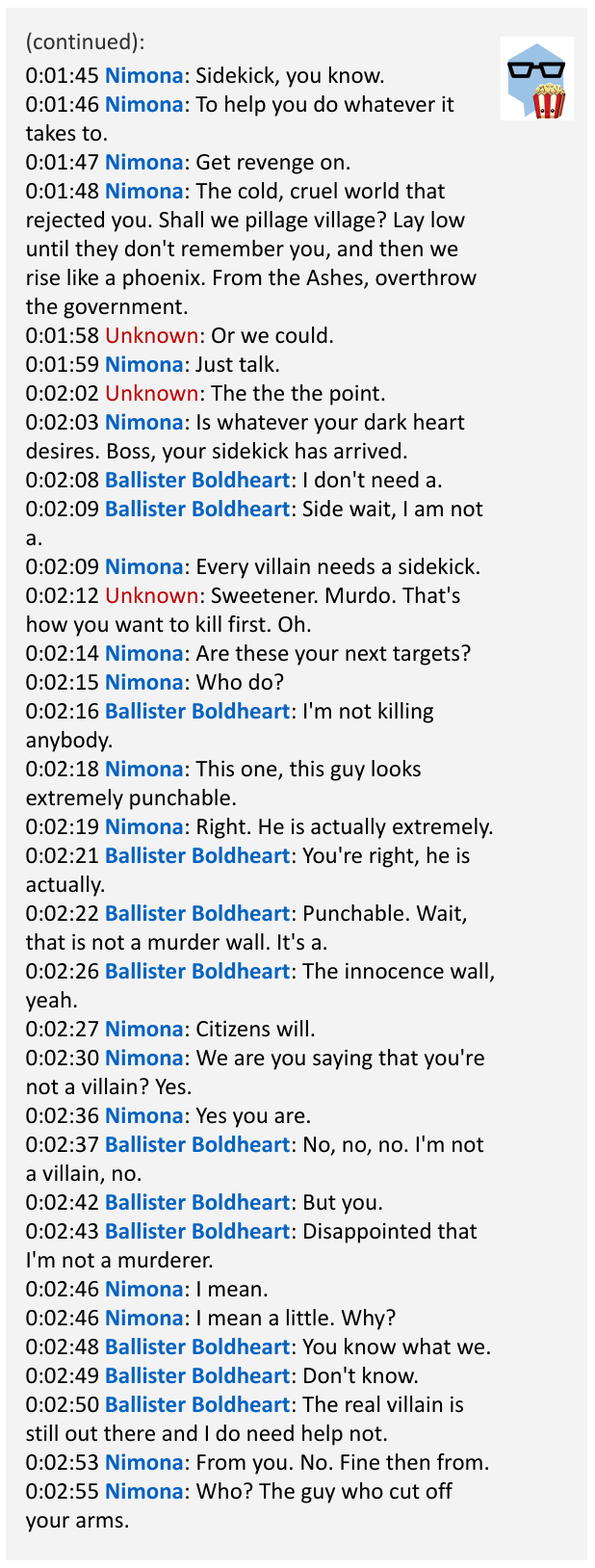}
\vspace{-1ex}
\caption{Case studies of \modelname’s capabilities and application scenarios: \textbf{speaker identification}; By leveraging visual prompting, \modelname can enhance ASR predictions with the speakers' identity. \bluehighlight~and~\redhighlight~highlight the correct and incorrect predictions, respectively. The original video is available at \url{https://www.netflix.com/title/81444554}, from 9:52 to 12:52.
}
\label{fig:vid-speaker-id-2}
\end{figure*}

\Paragraph{Speaker Identification.} Our exploration has revealed another valuable application of visual prompting in enhancing the quality of Automatic Speech Recognition (ASR). In Figures~\ref{fig:vid-speaker-id-1}-\ref{fig:vid-speaker-id-2}, we highlight a scenario where conventional ASR struggles to accurately recognize the number of speakers and their identities in the video. Visual prompting plays a pivotal role in enhancing ASR performance by providing contextual cues to identify individuals and attribute speech to specific speakers. This improvement ensures more precise transcriptions, enabling a more accurate representation of the dialogue and interactions within the video content.

\begin{figure*}[th]
\centering
\vspace{-40pt}
\includegraphics[width=1.0\textwidth]{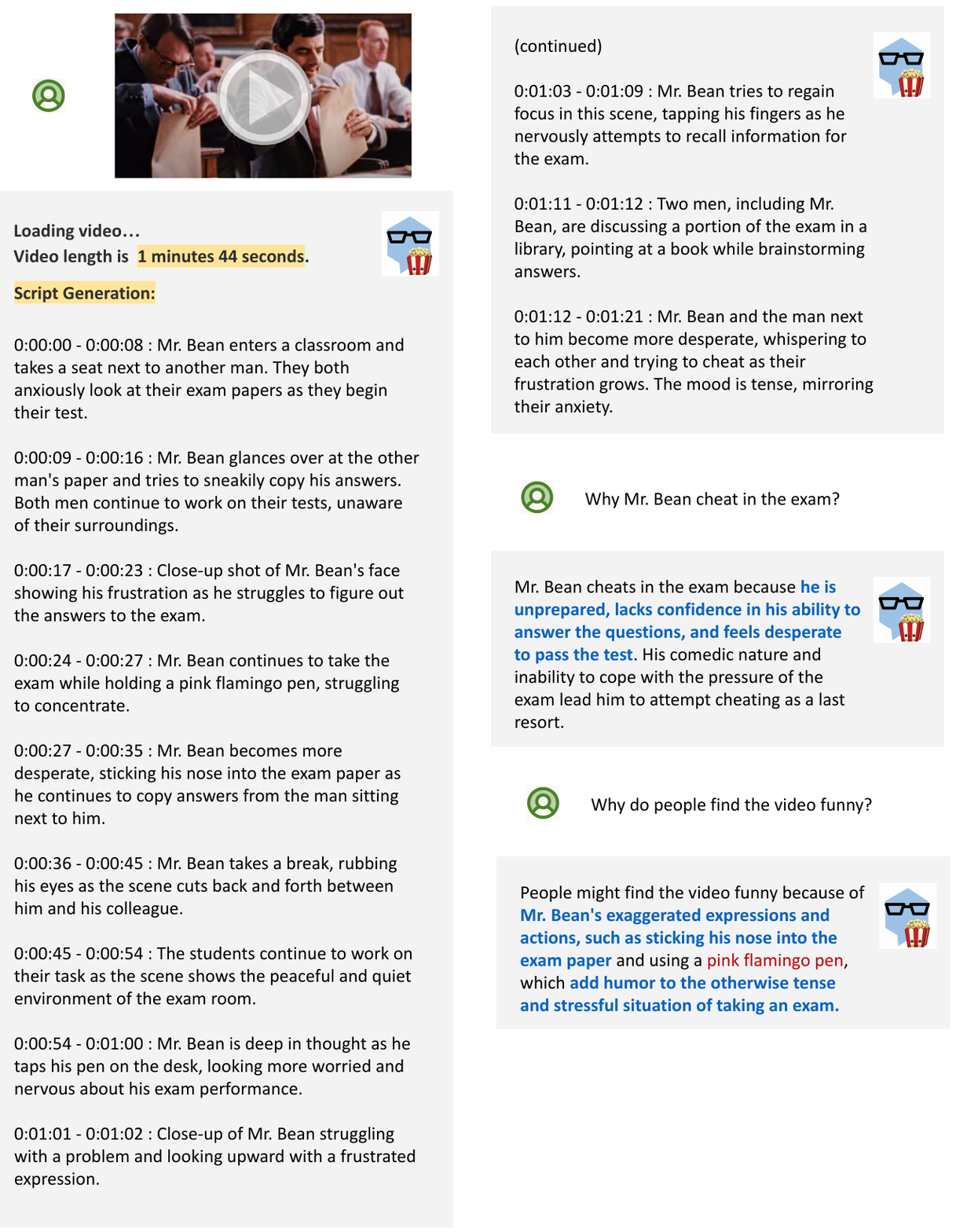}
\vspace{-1ex}
\caption{Case studies of \modelname’s capabilities and application scenarios: \textbf{audio description generation}; Without ASR inputs, \modelname processes the video and generates a detailed script. \bluehighlight~and~\redhighlight~highlight the correct and incorrect predictions, respectively. The original video is available at \url{https://www.youtube.com/watch?v=29ZmXkICT1g}, from 2:50 to 4:34.
}
\label{fig:vid-mrbean}
\end{figure*}

\Paragraph{Audio Description Generation.} Audio descriptions~\cite{rohrbach2015dataset, han2023autoad} play a crucial role in making videos accessible to individuals who are blind, have low vision, or face difficulties in visually understanding the content. These descriptions provide contextual narration of meaningful visual elements, clarify speakers, and convey the essence of visual information within a video. In our experiments, we also explore \modelname's performance in audio description generation. We experiment with videos where there is limited or no speech content. In Figure~\ref{fig:vid-mrbean}, we showcase an example featuring a short film of Mr. Bean taking an exam, which primarily lacks speech. Without ASR inputs, \modelname processes the video and generates a detailed script. This shows \modelname's versatility in handling various types of video content and its potential in creating inclusive and accessible multimedia content.

\begin{figure*}[th]
\centering
\vspace{-40pt}
\includegraphics[width=1.0\textwidth]{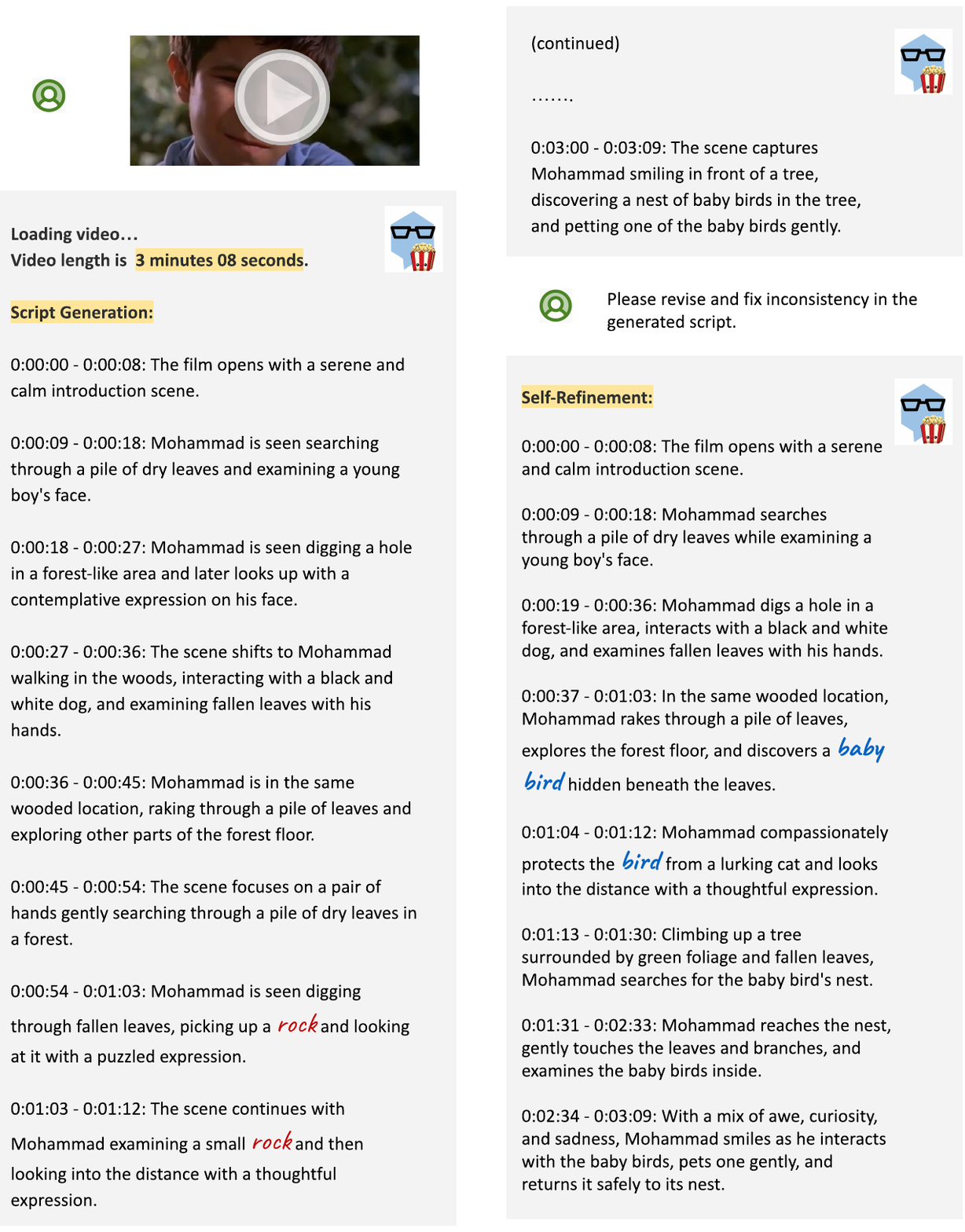}
\vspace{-1ex}
\caption{Case studies of \modelname’s capabilities and application scenarios: \textbf{self-refinement}; In this example, \modelname mistakenly identifies a bird as a rock due to low-resolution video frames. The self-refinement helps rectify the errors, resulting in a more refined output. \bluehighlight~and~\redhighlight~highlight the correct and incorrect predictions, respectively. The original video is available at \url{https://www.youtube.com/watch?v=o4UKvgBkO2Y}, from 1:06 to 4:17.
}
\label{fig:vid-refine}
\end{figure*}

\Paragraph{Self-Refinement.} While the generated script offers a comprehensive understanding of video content, our experiments have unveiled occasional inaccuracies, especially in cases involving blurry or low-resolution video frames, as demonstrated in Figure~\ref{fig:vid-refine}. In this example, \modelname mistakenly identifies a bird as a rock due to the challenges posed by the video's visual quality. To address such inconsistencies and elevate the overall accuracy of the generated script, we employ a self-refinement approach~\cite{madaan2023self,shinn2023reflexion,yang2023idea2img}. This involves revising the script based on both the initially generated script and a concurrently generated video summary. Through this process, \modelname is able to rectify errors and inaccuracies, resulting in a more refined output. 

\begin{figure*}[th]
\centering
\vspace{-40pt}
\includegraphics[width=1.0\textwidth]{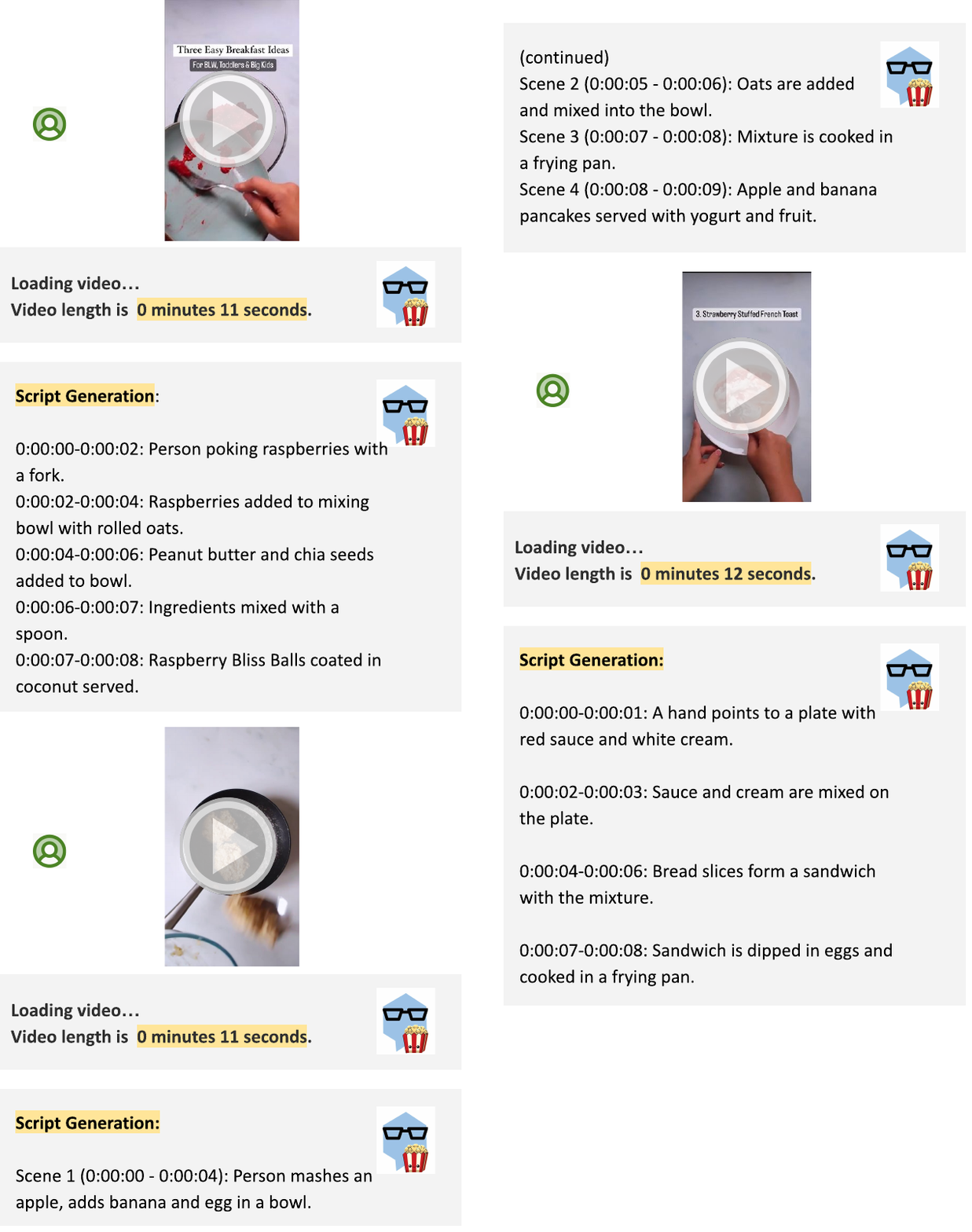}
\vspace{-1ex}
\caption{Case studies of \modelname’s capabilities and application scenarios: \textbf{fast-changing short videos.} The original videos are available at \url{https://www.instagram.com/mealtimewithmummy/reels/}
}
\label{fig:vid-tiktok}
\end{figure*}

\Paragraph{Fast-Changing Short Videos.} In Figure~\ref{fig:vid-tiktok}, we present an example of our experimentation with fast-changing short-form videos, such as those found on platforms like TikTok. Short videos often feature non-standard frame sizes and significantly shorter durations compared to conventional videos. Remarkably, \modelname excels at accurately describing the cooking recipes depicted in these short videos, despite the distinct characteristics of such content.

These examples demonstrate the versatility of \modelname in processing a diverse array of video content. Whether dealing with lengthy documentaries, episodic series, or short-form clips, \modelname adapts seamlessly to the unique attributes of each video type, consistently delivering meaningful and contextually relevant descriptions. 

\subsection{Applications to Interactive Environments}

In the following section, we evaluate \modelname when applying to the context of streaming inputs. \modelname serves as an agent in an interactive environment, continually receiving streaming video frames as the inputs.

\begin{figure*}[th]
\centering
\vspace{-40pt}
\includegraphics[width=1.0\textwidth]{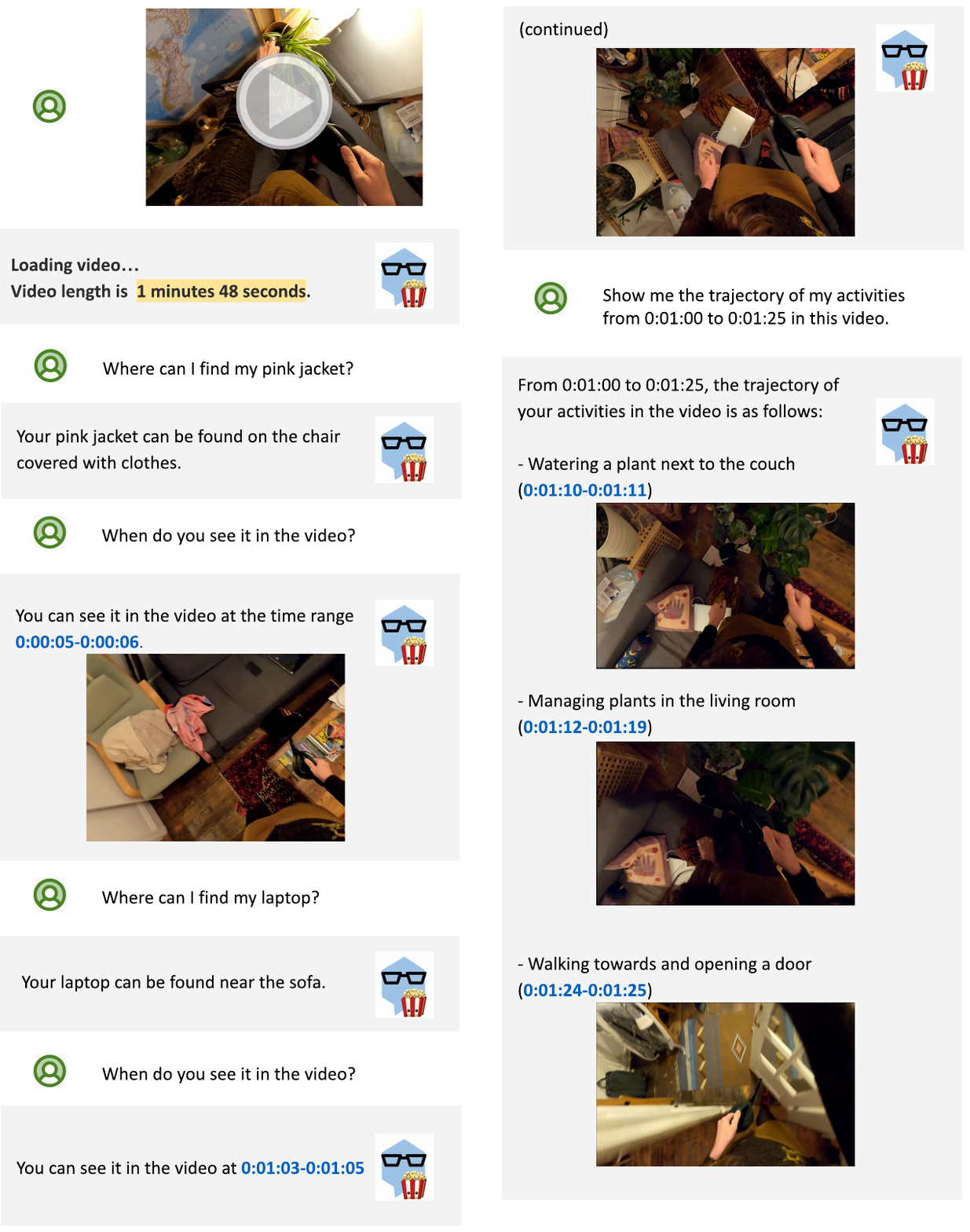}
\vspace{-1ex}
\caption{Case studies of \modelname’s capabilities and application scenarios: \textbf{embodied agent}. \bluehighlight~highlights the correct prediction. The original video is collected from Ego4D dataset~\cite{grauman2022ego4d}.
}
\label{fig:vid-embodied}
\end{figure*}

\Paragraph{Embodied Agent.} Figure~\ref{fig:vid-embodied} illustrates an example where \modelname is applied to an egocentric video captured by a head-mounted camera. This video, collected from Ego4D dataset~\cite{grauman2022ego4d}, provides a brief glimpse into the wearer's daily life within their home environment. Remarkably, \modelname showcases its capability in understanding such video content and assists the user in a few practical tasks. Specifically, \modelname helps the user locate items like the pink jacket and the laptop within the home. Additionally, it generates a list of the user's activities within a specified time range, offering insights into the wearer's daily routine. 

\begin{figure*}[th]
\centering
\vspace{-40pt}
\includegraphics[width=1.0\textwidth]{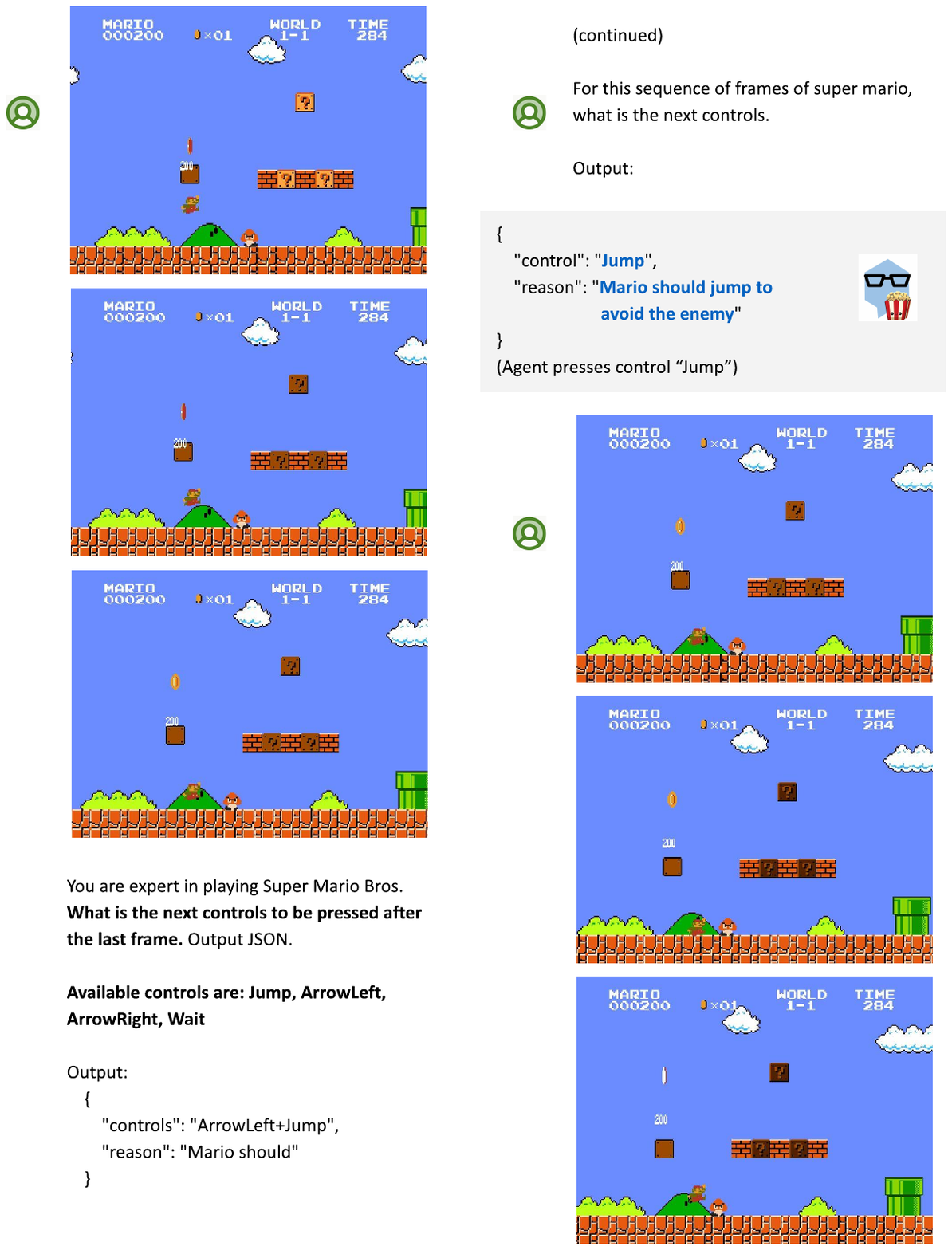}
\vspace{-1ex}
\caption{Case studies of \modelname’s capabilities and application scenarios: \textbf{playing video game.} \bluehighlight~highlights the correct prediction. Figures \ref{fig:vid-game-2}-\ref{fig:vid-game-4} show continued outputs. The video is generated by Pygame library~\cite{pygame}. 
}
\label{fig:vid-game}
\end{figure*}

\begin{figure*}[th]
\centering
\vspace{-40pt}
\includegraphics[width=1.0\textwidth]{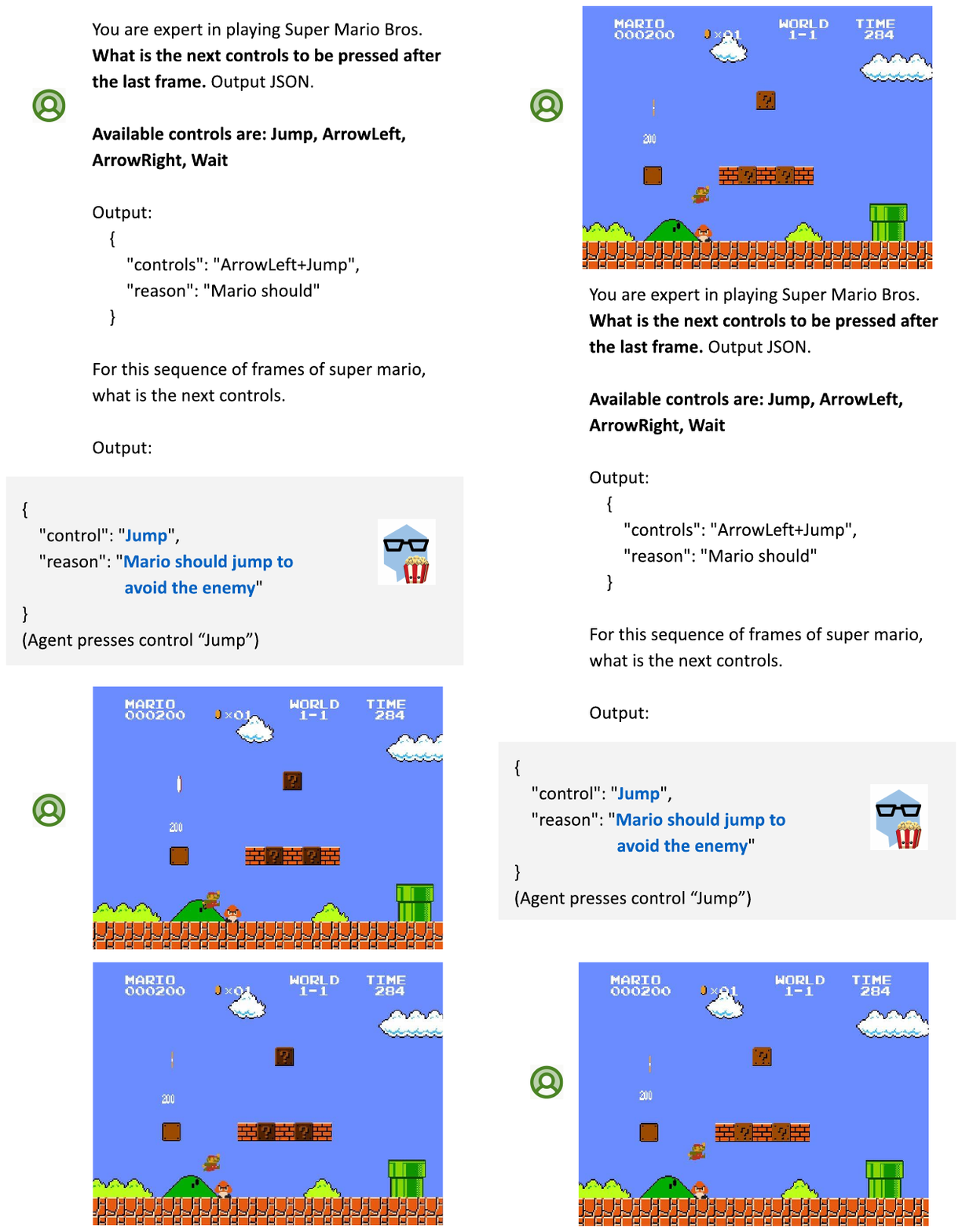}
\vspace{-1ex}
\caption{Case studies of \modelname’s capabilities and application scenarios: \textbf{playing video game.} \bluehighlight~highlights the correct prediction. Figures \ref{fig:vid-game-3}-\ref{fig:vid-game-4} show continued outputs. The video is generated by Pygame library~\cite{pygame}. 
}
\label{fig:vid-game-2}
\end{figure*}

\begin{figure*}[th]
\centering
\vspace{-40pt}
\includegraphics[width=1.0\textwidth]{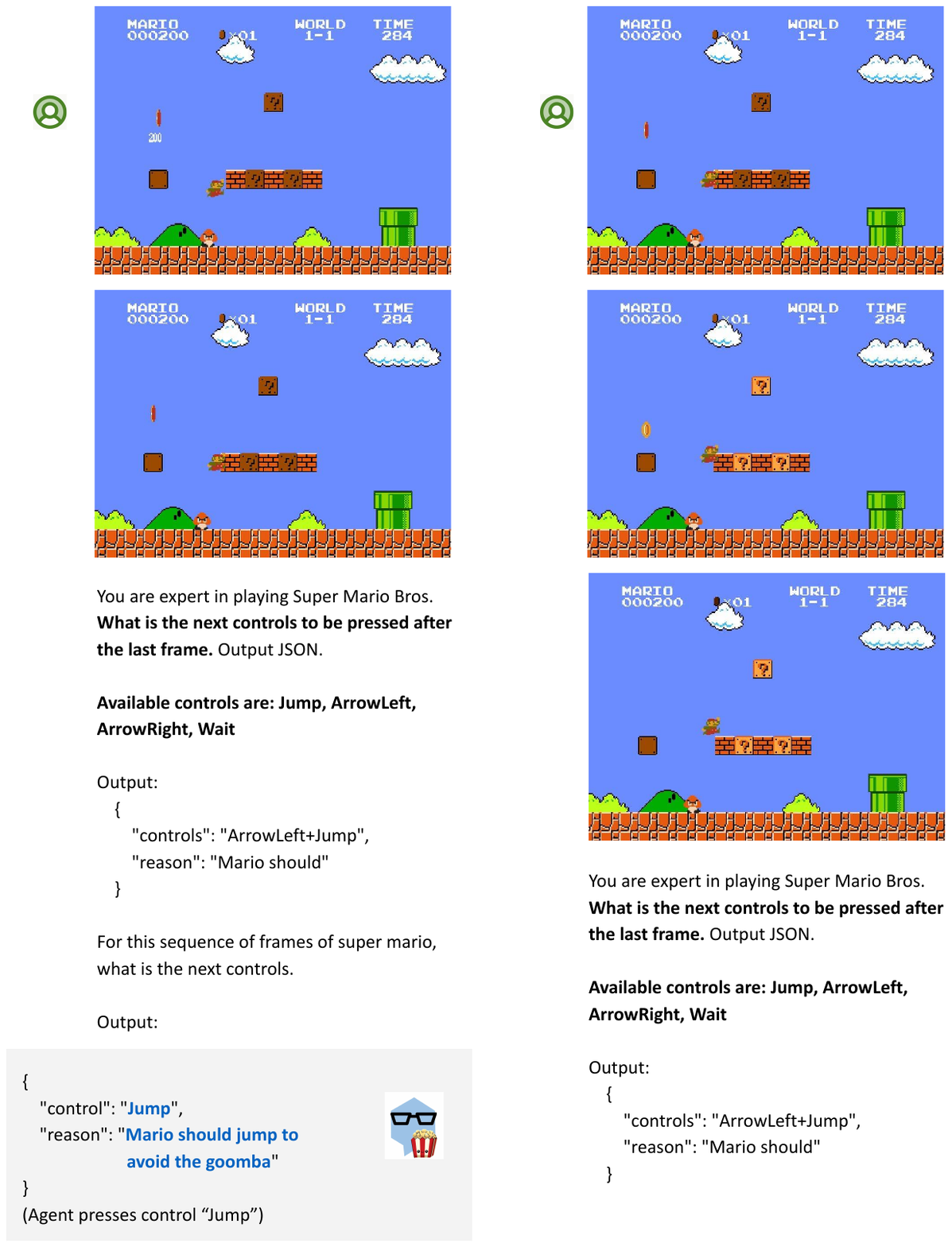}
\vspace{-1ex}
\caption{Case studies of \modelname’s capabilities and application scenarios: \textbf{playing video game.} \bluehighlight~highlights the correct prediction. Figure \ref{fig:vid-game-4} shows continued outputs. The video is generated by Pygame library~\cite{pygame}. 
}
\label{fig:vid-game-3}
\end{figure*}

\begin{figure*}[th]
\centering
\vspace{-40pt}
\includegraphics[width=1.0\textwidth]{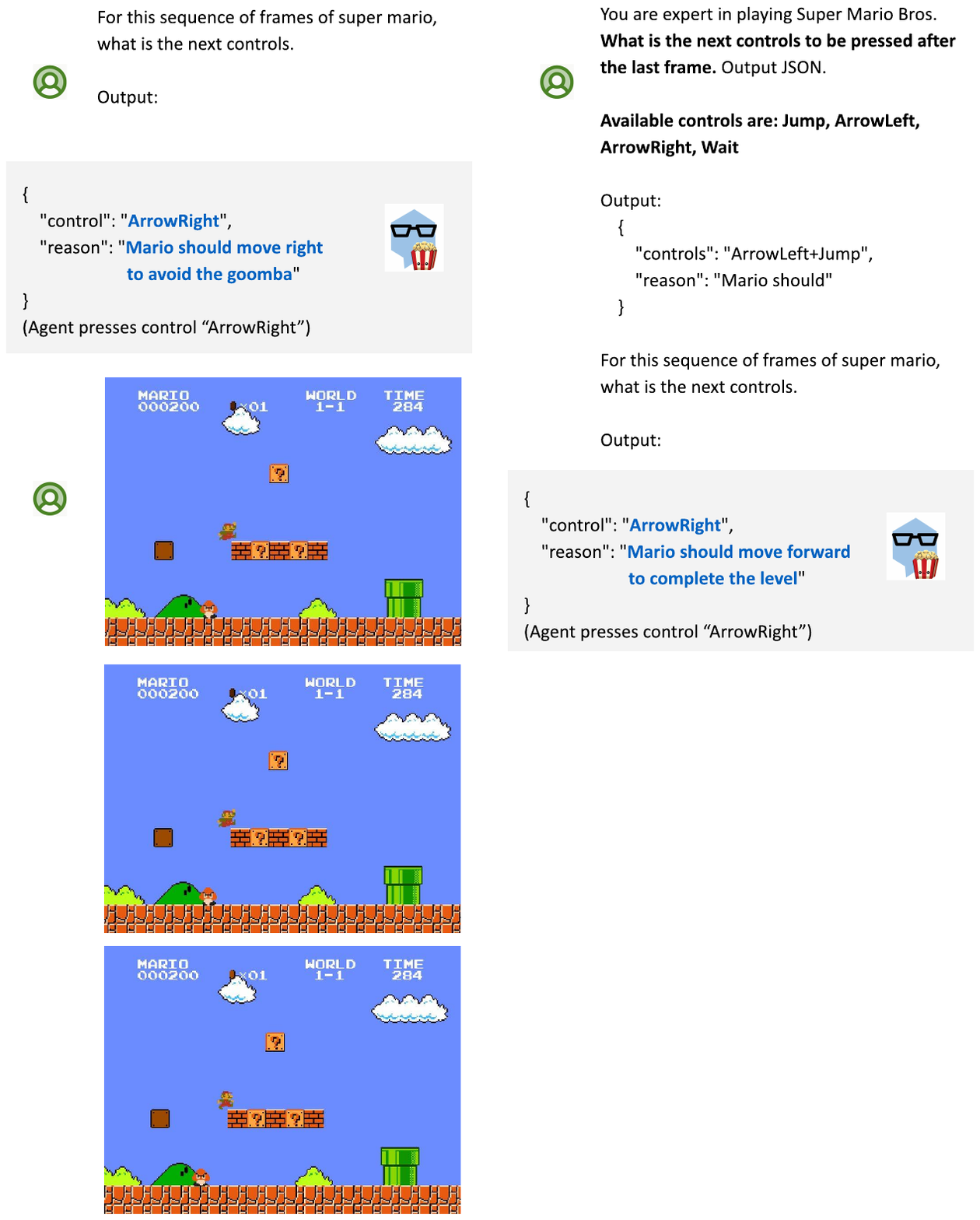}
\vspace{-1ex}
\caption{Case studies of \modelname’s capabilities and application scenarios: \textbf{playing video game.} \bluehighlight~highlights the correct prediction. The video is generated by Pygame library~\cite{pygame}. 
}
\label{fig:vid-game-4}
\end{figure*}

\Paragraph{Playing Video Games.} Figures~\ref{fig:vid-game}-\ref{fig:vid-game-4} demonstrate the results of applying \modelname to a Mario video game~\cite{pygame}. In these experiments, our agent consistently receives three video frames as states and calculates the next possible control action. Remarkably, our agent displays an understanding of the specific video game dynamics and generates reasonable action controls to play the game effectively. These examples highlight \modelname's ability to comprehend and navigate in an interactive gaming environment. Interested readers may find the full gameplay demonstration on our project website.

\begin{figure*}[th]
\centering
\vspace{-40pt}
\includegraphics[width=1.0\textwidth]{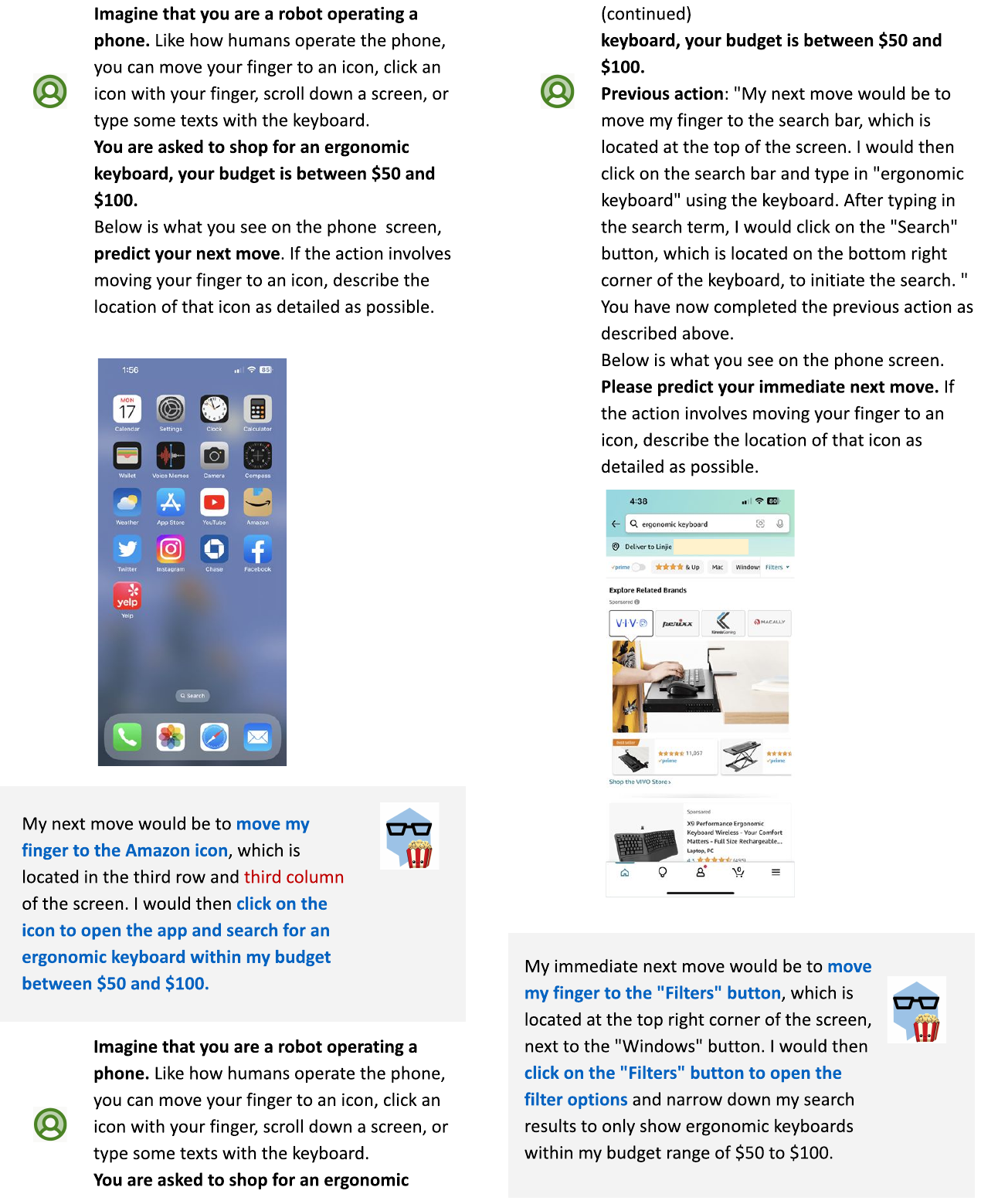}
\vspace{-1ex}
\caption{Case studies of \modelname’s capabilities and application scenarios: \textbf{GUI navigation.} \bluehighlight~and~\redhighlight~highlight the correct and incorrect predictions, respectively. Figures \ref{fig:vid-gui-2}-\ref{fig:vid-gui-5} show continued outputs. 
}
\label{fig:vid-gui-1}
\end{figure*}

\begin{figure*}[th]
\centering
\vspace{-40pt}
\includegraphics[width=1.0\textwidth]{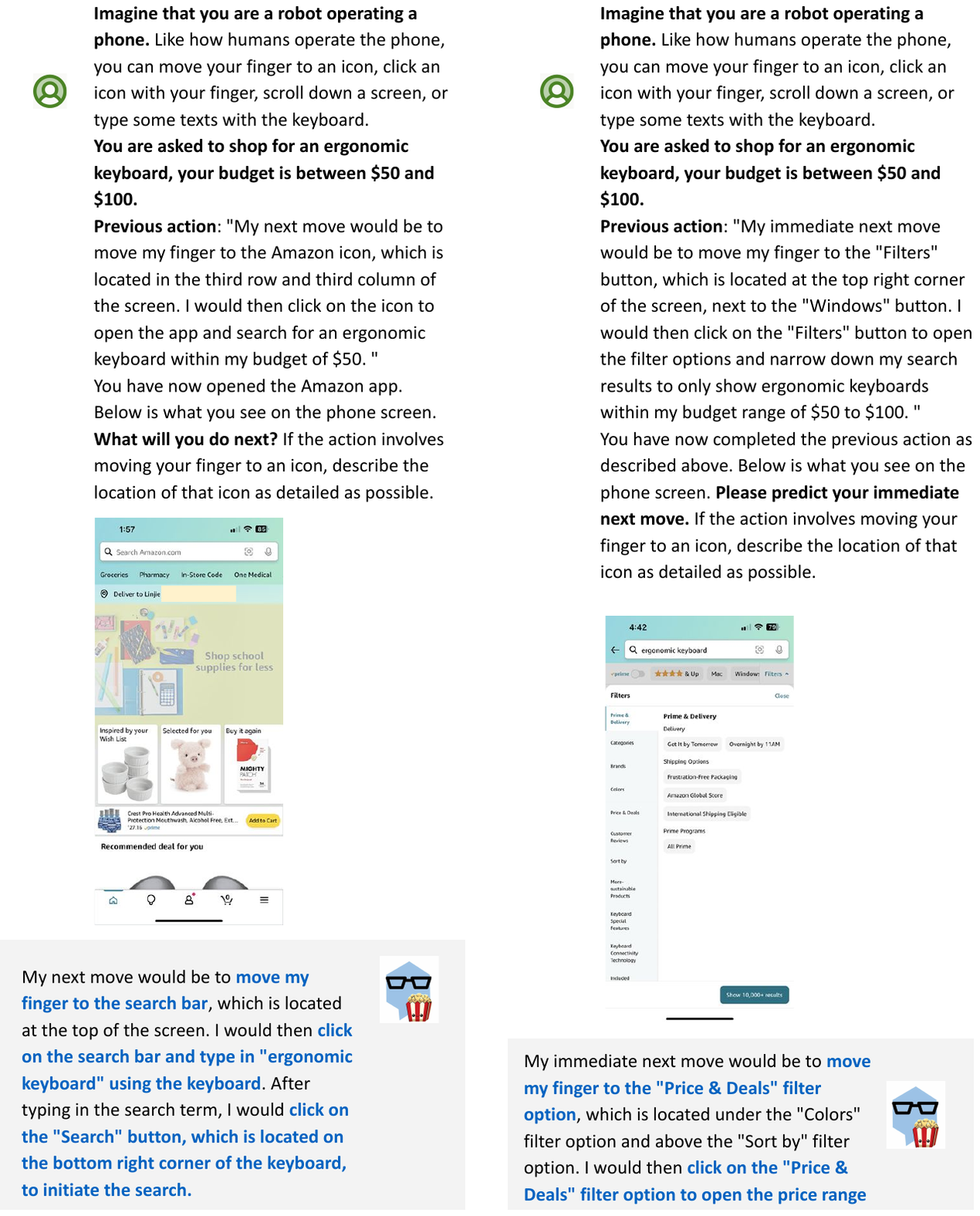}
\vspace{-1ex}
\caption{Case studies of \modelname’s capabilities and application scenarios: \textbf{GUI navigation.} \bluehighlight~and~\redhighlight~highlight the correct and incorrect predictions, respectively. Figures \ref{fig:vid-gui-3}-\ref{fig:vid-gui-5} show continued outputs. 
}
\label{fig:vid-gui-2}
\end{figure*}

\begin{figure*}[th]
\centering
\vspace{-40pt}
\includegraphics[width=1.0\textwidth]{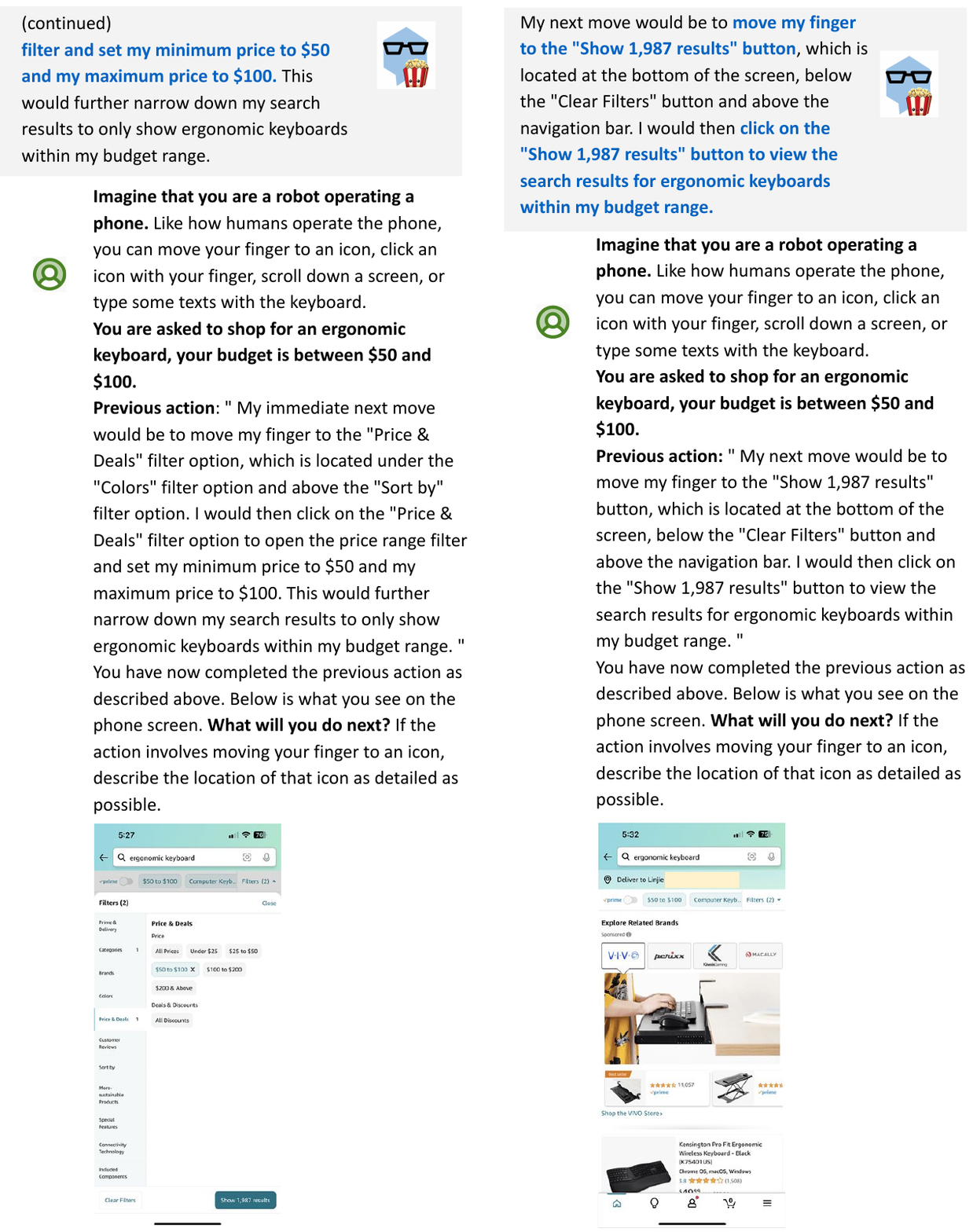}
\vspace{-1ex}
\caption{Case studies of \modelname’s capabilities and application scenarios: \textbf{GUI navigation.} \bluehighlight~and~\redhighlight~highlight the correct and incorrect predictions, respectively. Figures \ref{fig:vid-gui-4}-\ref{fig:vid-gui-5} show continued outputs. 
}
\label{fig:vid-gui-3}
\end{figure*}

\begin{figure*}[th]
\centering
\vspace{-40pt}
\includegraphics[width=1.0\textwidth]{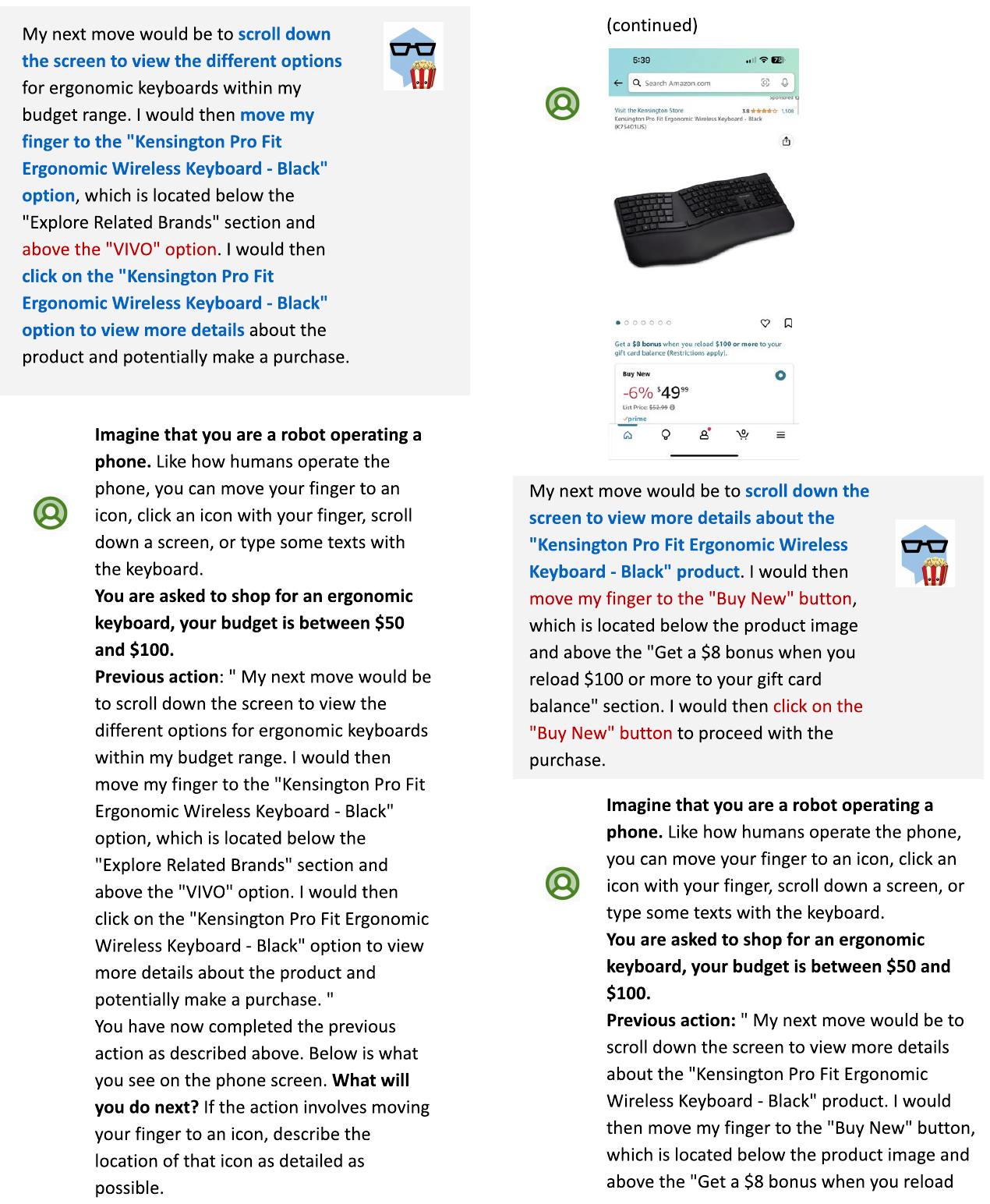}
\vspace{-1ex}
\caption{Case studies of \modelname’s capabilities and application scenarios: \textbf{GUI navigation.} \bluehighlight~and~\redhighlight~highlight the correct and incorrect predictions, respectively. Figure \ref{fig:vid-gui-5} shows continued outputs. 
}
\label{fig:vid-gui-4}
\end{figure*}

\begin{figure*}[th]
\centering
\vspace{-40pt}
\includegraphics[width=1.0\textwidth]{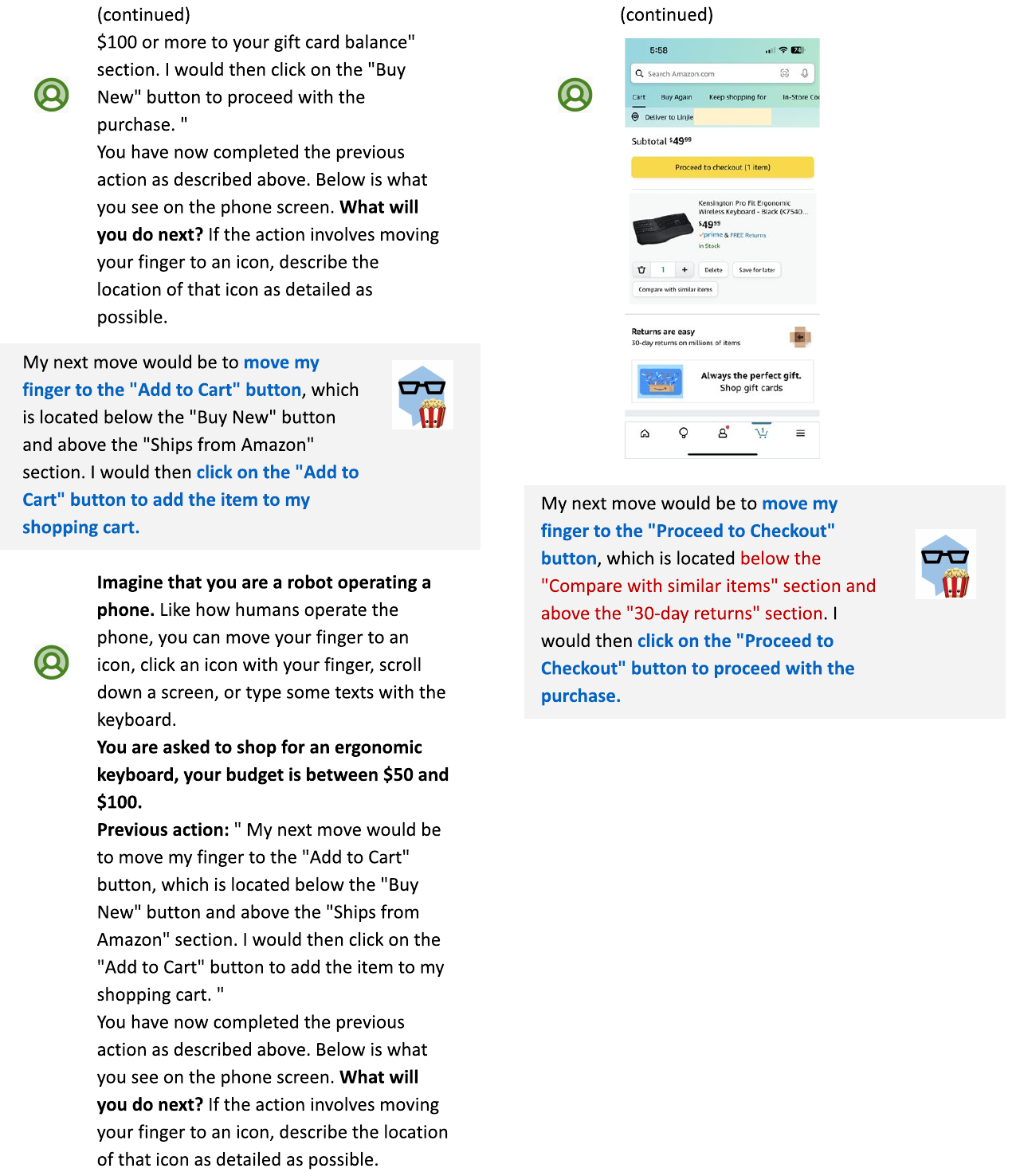}
\vspace{-1ex}
\caption{Case studies of \modelname’s capabilities and application scenarios: \textbf{GUI navigation.} \bluehighlight~and~\redhighlight~highlight the correct and incorrect predictions, respectively.
}
\label{fig:vid-gui-5}
\end{figure*}

\Paragraph{GUI Navigation.} Figures~\ref{fig:vid-gui-1}-\ref{fig:vid-gui-5} provide the demonstration of \modelname's performance in the GUI navigation scenario. In this context, the agent continually receives iPhone screenshots and previous user actions as states. The agent effectively predicts the possible next steps in the user's journey, which may include clicking on the correct shopping apps, initiating searches for items of interest, and ultimately placing an order. These results demonstrate \modelname's remarkable ability to interact with graphical user interfaces, facilitating seamless and intelligent navigation through digital interfaces.

\begin{table}[t]
\caption{Questionnaire for the group with visual impairments. Participants listen to a video and subsequently assign scores (ranging from 0 to 10) for distinct auditory criteria.
}
\label{tab:blind}
\vspace{-4mm}
\begin{tcolorbox} 
    \small
    \begin{tabular}{p{0.99\columnwidth}}

\textbf{Effectiveness of Delivery:} If the original audio and the embedded AD are effectively presented?

\textbf{Informative:} Is it easy to follow the storyline? Does the AD provide context and background information when necessary?

\textbf{Audio Quality:} Is the overall audio production quality good?

\textbf{Overall Satisfaction:} Are you satisfied with the overall AD experience?
    \end{tabular}
\end{tcolorbox}
\end{table}

\begin{table}[t]
\caption{
Questionnaire for the group with normal vision. Participants view a video and subsequently assign scores (ranging from 0 to 10) for various auditory and visual criteria.
}
\label{tab:sight}
\vspace{-4mm}
\centering
\begin{tcolorbox} 
    \centering
    \small
    \begin{tabular}{p{0.99\columnwidth}}

\textbf{Clarity:} Are the visual elements clearly and accuratetly described?

\textbf{Conciseness:} Does the AD convey essential visual information without overloading the user?

\textbf{Timing and Synchronization:} Are the original audio and the embedded AD effectively presented? Does the AD properly synchronize with visual contents?

\textbf{Informative:} Is it easy to follow the storyline? Does the AD provide context and background information when necessary?

\textbf{Audio Quality:} Is the overall audio production quality good?

\textbf{Overall Satisfaction:} Are you satisfied with the overall AD experience?
    \end{tabular}
\end{tcolorbox}
\end{table}

\subsection{User Study}

We explore the potential of \modelname for people who are blind or have low vision. Audio description (AD)~\cite{rohrbach2015dataset, han2023autoad} provides an auditory narration integrated into the video's soundtrack, offering important visual details that may not be discernible from the main video soundtrack. Such descriptions play a pivotal role in conveying essential visual content to those with visual impairments.

To assess the efficacy of \modelname in generating audio descriptions (AD), we conduct a user study. We invited 9 participants for the evaluation. 4 participants were either blind or had low vision, while the remaining 5 had normal vision. All the participants have normal hearing. For the purposes of the experiments, we segregated participants into two distinct groups: (\textit{i}) Group with visual impairments, and (\textit{ii}) Group with normal vision.

\subsubsection{Evaluation Procedure}

Our experiments utilize a curated set of videos, which are mainly suggested by the American Council of the Blind\footnote{The Audio Description Project: \url{https://adp.acb.org/}}. We also collected accessibility videos from YouTube\footnote{Apple Accessibility: \url{https://www.youtube.com/watch?v=SL7YSqlEd8k}}. For every video used in our evaluation, participants are exposed to two versions: the first containing human-crafted AD and the second powered by \modelname-generated AD. Both renditions are narrated using text-to-speech (TTS) technology.

We have designed two questionnaires for the two groups, referenced in Table~\ref{tab:blind} and Table~\ref{tab:sight}, respectively. Participants with visual impairments are instructed to base their evaluation exclusively on auditory cues. In contrast, those with normal vision are instructed to consider both visual and auditory elements.

The assessment adopts the standardized Likert scale for ratings. For each posed question, participants are guided to assign a score ranging from 0 to 10, with higher values indicating more favorable feedback. Furthermore, participants are urged to share feedback and remarks concerning their overall experience.

\subsubsection{Results on the Group with Visual Impairments}

\begin{figure}[t]
\centering
\includegraphics[width=1.0\columnwidth]{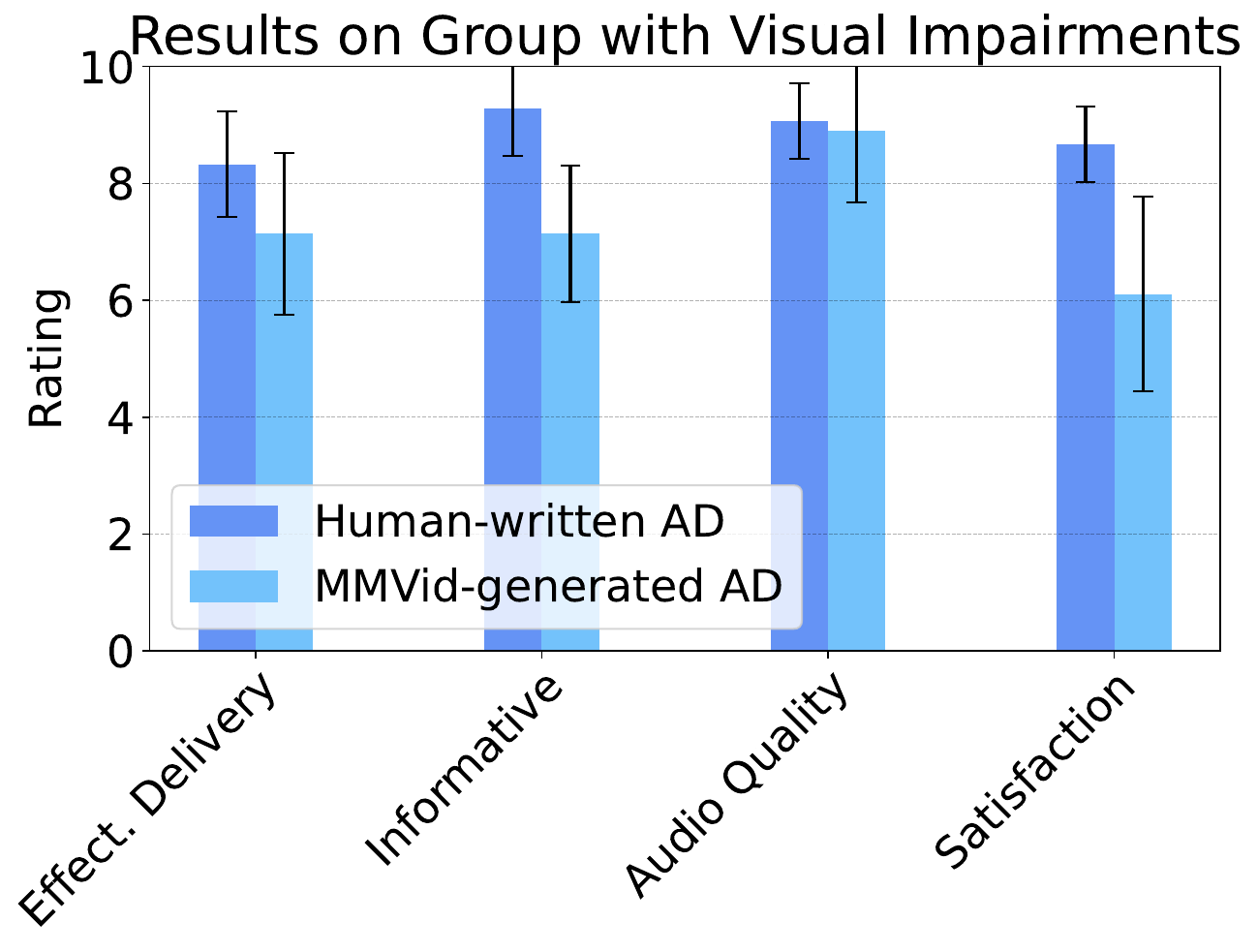}
\vspace{-4ex}
\caption{Results on the group with visual impairments. \modelname-generated AD is close to human-generated ones in terms of audio quality and effectiveness of delivery. However, \modelname's AD yields lower satisfaction levels compared to the human-generated ones. This was primarily attributed to occasional overlaps between the audio descriptions and the video dialogues.}
\label{fig:blind-results}
\end{figure}

We utilized 3 different videos for our evaluation, with durations of 1 minute, 1 minute 42 seconds, and 2 minutes 42 seconds, respectively.
Each of the 4 participants with visual impairment was well versed with screen reader and other common accessibility tools.  
After listening to the audio descriptions for each video, they were asked to respond to the 4 questions outlined in Table~\ref{tab:blind}.
\newline
\newline
\textbf{Hypotheses and Results}
\newline
\newline
\textbf{H1:} The \modelname-generated audio description and original video dialogues are effectively presented to the participants.
\newline
\textbf{Results:} Using the Likert scale (0=Not Effective to 10=Most Effective) the participants rated the effectiveness of the delivery of human-crafted AD and \modelname-generated AD. On average, participants gave $7.14 \pm 1.39$ for \modelname-generated AD and $8.33 \pm 0.90$ for human-crafted AD, which shows a \modelname-generated AD very close to human-crafted one in terms of effective delivery (Figure \ref{fig:blind-results}).  
\newline
\newline
\textbf{H2:} Participants are able to follow the main story line of the video based on \modelname-generated audio description only.
\newline
\textbf{Results:} Using the Likert scale (0=Not Informative to 10=Highly Informative) the participants rated the informativeness of human-crafted AD and \modelname-generated AD. On average, participants gave $7.14 \pm 1.16$ for \modelname-generated AD and $9.29 \pm 0.91$ for human-crafted AD, which shows little difference in informativeness between \modelname-generated AD and human-crafted one (Figure \ref{fig:blind-results}).
\newline
\newline
\textbf{H3:} \modelname-generated AD and human-crafted AD are close in terms of voice and audio quality. 
\newline
\textbf{Results:} Using the Likert scale (0=Low Quality to 10=High Quality) the participants rated the voice and audio quality on average as $8.91 \pm 1.23$ for \modelname-generated AD and $9.07 \pm 0.65$ for human-crafted AD. This minimal difference between the scores indicates the close-to-human voice and audio quality of \modelname-generated AD (Figure \ref{fig:blind-results}).
\newline
\newline
\textbf{Discussion:}  
\newline
The results show that the participants' overall satisfaction of \modelname-generated ADs was on average around 2 points less than human-crafted ones in the Likert scale (0=Not Satisfied to 10=Highly satisfied) (Figure \ref{fig:blind-results}). Some of the difficulties indicated by participants while listening to \modelname-generated ADs were \textbf{1)} occasional overlaps between AD audio and original video dialogues \textbf{2)} wrong descriptions due to hallucinations of \gptmodelnamefull. Regardless of the difference in overall satisfaction, all the participants agreed that \modelname-generated AD can provide a cost-effective and scalable solution. Thus, millions of videos that cannot afford to be professionally audio described, can be auto-processed by a tool like \modelname to make them accessible to the visual-impaired community.

\subsubsection{Results on the Group with Normal Vision}
\begin{figure}[t]
\centering
\includegraphics[width=1.0\columnwidth]{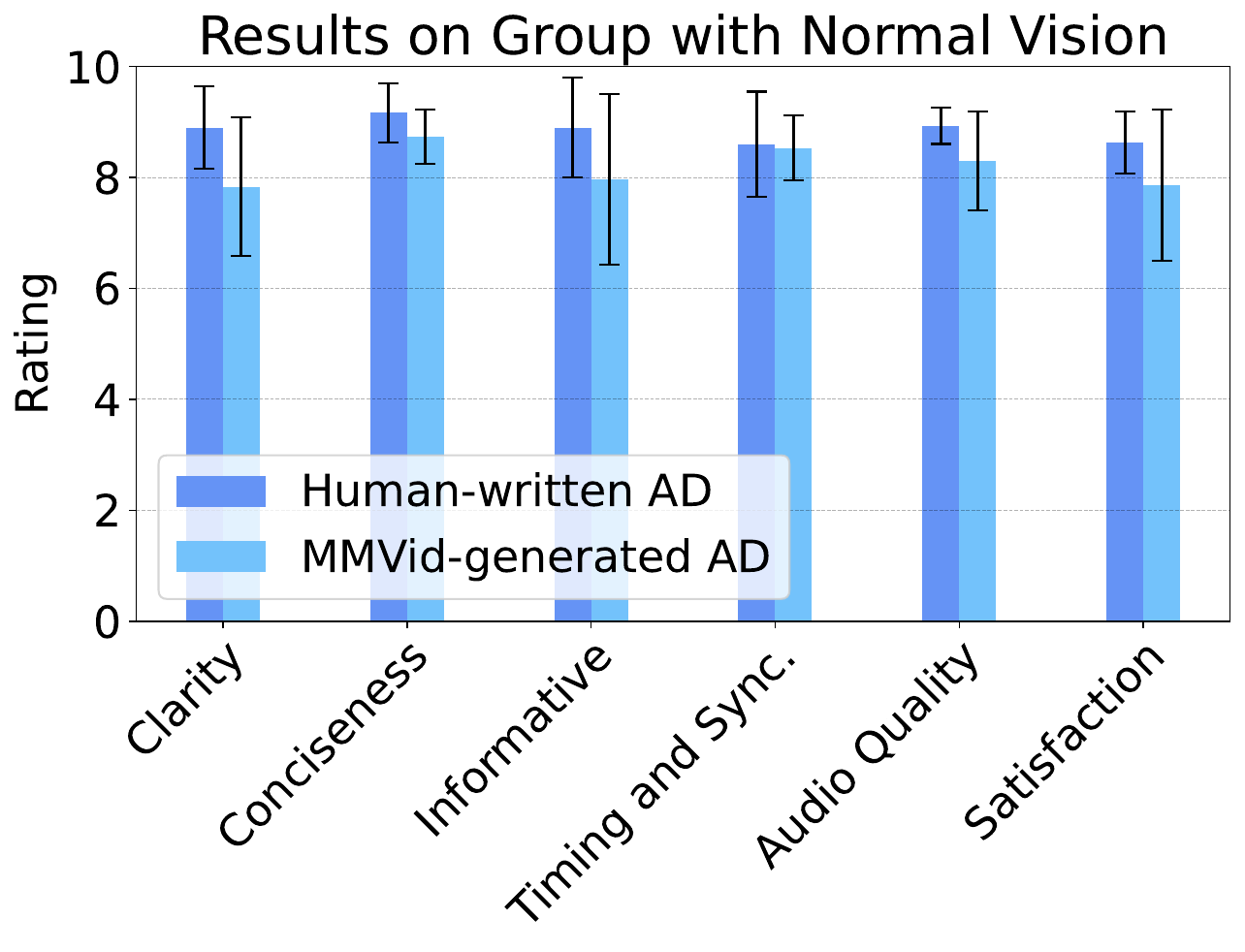}
\vspace{-4ex}
\caption{Results on the group with normal vision. \modelname-generated AD was on average $<$1 points lower than human-generated ADs. The participants were able to follow the main story line and the audios are in sync with the video content.}
\label{fig:sight-results}
\end{figure}

For sighted individuals, we used the same set of videos as we used for individuals with visual impairments. All of our 5 participants answered to 6 questions listed in Table~\ref{tab:sight} after watching videos embedded with \modelname-generated AD as subtitles and audio track.  
\newline
\newline
\textbf{Hypotheses and Results}
\newline
\newline
\textbf{H1:} The \modelname-generated AD is accurate and conveys essential information without overloading the listener.
\newline
\textbf{Results:} The sighted individuals rated the clarify and accuracy of \modelname-generated AD as $7.83 \pm 1.24$ and human-curated AD as $8.9 \pm 0.74$ on average, using the Likert scale (0=Not Accurate to 10=Most Accurate). In terms of conciseness, the participants on average gave $8.73 \pm 0.49$ for the \modelname-generated AD and $9.16 \pm 0.54$ for human-curated AD based on the Likert scale (0=Not concise to 10=Most concise). These results indicate \modelname-generated ADs are close to human-curated ones in terms of accuracy and conciseness (Figure \ref{fig:sight-results}).
\newline
\newline
\textbf{H2:} The \modelname-generated ADs are in sync with visual content and do not overlap with other dialogues ensuring listeners can follow the story line.
\newline
\textbf{Results:} Participants gave on average $8.90 \pm 0.90$ and $7.97 \pm 1.54$ to human-crafted AD and \modelname-generated AD respectively using the Likert scale (0=Not Informative to 10=Highly Informative). Human-crafted AD and \modelname-generated AD received $8.59 \pm 0.95$ and $8.53 \pm 0.58$ respectively on the aspect of timing and synchronization using the Likert scale (0=Not Effective to 10=Most Effective). These indicates while listening to \modelname-generated ADs participants were able to follow main story line and found the audios are in sync with video content very close to that of human-crafted ADs (Figure \ref{fig:sight-results}).
\newline
\newline
\textbf{H3:} The voice and audio quality of \modelname-generated ADs are close to human-crafted ADs.
\newline
\textbf{Results:} The results are very similar to results on group with visual impairments. Sighted participants rated the voice and audio quality on average as $8.30 \pm 0.89$ for \modelname-generated AD and as $8.93 \pm 0.32$ for human-crafted AD. Therefore the voice and audio experience did not degrade much while listening to \modelname-generated ADs compare to human-crafted ADs (Figure \ref{fig:sight-results}).
\newline
\newline
\textbf{Discussion:}  
\newline
The evaluations on sighted individuals helped to cross verify the hypotheses of individuals with visual impairments, that are based on audio cues only. Although the overall satisfaction points for sighted participants with \modelname-generated ADs was on average $<$1 points lower than human-generated ADs (Figure \ref{fig:sight-results}), the overall satisfaction points for participants who were blind was worse. This is expected because sighted individuals had access to both audio and video modalities but individuals with visual impairments did not. We also believe the reason for lower overall satisfaction, may have been the lack of practice listening to auto generated ADs. Some of the users also mentioned they have preference between pitches of voice and number of concurrent audio channels. These may add to the reason of lower overall satisfaction.

\subsubsection{Participant Feedback}

We present a collection of interview quotes from our participants who were visually impaired, in which they share their personal experiences and insights about the audio descriptions (AD) generated by \modelname. The participants expressed a unanimous desire to continue utilizing this AD generation service in the future, highlighting its exceptional quality (``Nearly perfect''), intricate details (``favorite was the details''), extensive applicability (``allowed me to follow anything visual''), and the profound impact it has on them (``I did not depend on someone else''). Below, we provide additional quotes for further insight.

\begin{quote}
P1: \textit{``I understand what is going on very quickly and I did not depend on someone else.''}\\
P2: \textit{``If it's AI-generated, there are so many places it's not available, and we need it there.''}\\
P2: \textit{``First time listening to auto-generated AD. As a user, if I am offered this AD, I would take it.''}\\
P3: \textit{``Nearly perfect. Most favorite was the details.''}\\
P3: \textit{``More information helped me follow the storyline.''}\\
P3: \textit{``It allowed me to follow anything visual. It felt natural the way AD describes how the actor interacts with the environment.''}\\
P3: \textit{``I love animal kingdom, and I watch Wild Earth safari virtual tour. I would love to have audio descriptions of Wild Earth videos and daily safaris.''}\\
P4: \textit{``I would like to have auto-generated audio description for live conferences in Microsoft Teams.''}\\
P4: \textit{``It worked best as the original audio had not much value.''}
\end{quote}

Despite the positive feedback, not all responses were favorable:

\begin{quote}
P4: \textit{``I am skeptical when it becomes subjective. Sometimes I feel they make up stories which is not good.''}\\
P4: \textit{``After listening to the human-generated AD, I figured I misunderstood parts of the original story.''}\\
P1: \textit{``It keeps referring to the same person using their names instead of pronouns.''}\\
P4: \textit{``I don't deal well with overlapped or two parallel audios.''}
\end{quote}

Interestingly, even those participants who provided critical feedback still rated the \modelname-generated AD closely to human-generated AD, during the questionnaire sessions. This indicates that, similar to human-curated AD, adapting to \modelname-generated ADs might necessitate some practice and acclimatization over time.

\section{Conclusion}
We have presented \modelname, a system that synergizes with \gptmodelname for advancing video understanding. \modelname employs \gptmodelname to transcribe video content into long and detailed scripts, thereby enriching LLMs with advanced video understanding capabilities. 
Experimental results demonstrate the effectiveness of \modelname in addressing challenging tasks, including comprehension of hour-long videos, analysis across multiple episodes, identification of characters and speakers, and interaction with video games and graphical user interfaces.

Beyond the development of the \modelname system, we conducted an extensive user study, drawing feedback from a varied group of participants. The outcomes of this study indicated that the audio descriptions generated by \modelname closely mirror the quality of those crafted by humans. In our future work, we plan to explore SoM~\cite{yang2023set} and object tracking techniques to enhance various tasks and functionalities.

\subsection*{Acknowledgment}
\vspace{-2pt}

We are deeply grateful to OpenAI for providing access to their exceptional tool~\cite{gpt4,gpt4v,gpt4vcontribution,gpt4vblog}. We are profoundly thankful to Misha Bilenko for his invaluable guidance and support. We also extend heartfelt thanks to our Microsoft colleagues for their insights, with special acknowledgment to Cenyu Zhang, Saqib Shaikh, Ailsa Leen, Jeremy Curry, Crystal Jones, Roberto Perez, Ryan Shugart, Anne Taylor for their constructive feedback.

{\small
\bibliographystyle{ieee_fullname}
\bibliography{egbib, video}
}
\clearpage


\end{document}